\documentclass[10pt,twocolumn,letterpaper]{article}

\usepackage{cvpr}              

\usepackage{graphicx}
\usepackage{amsmath}
\usepackage{amssymb}
\usepackage{booktabs}
\usepackage{subfiles}
\usepackage{boldline}
\usepackage{multirow}
\usepackage{amsbsy}
\usepackage{color, colortbl}
\usepackage{xcolor}

\definecolor{purple}{rgb}{1,0,1}

\usepackage[T1]{fontenc}
\usepackage[utf8]{inputenc}
\usepackage{pict2e,picture}
\newsavebox\CBox
\newlength\CLength

\usepackage[pagebackref,breaklinks,colorlinks]{hyperref}

\newcount\Comments 
\usepackage{array,multirow,graphicx}
\usepackage{color}
\usepackage{wrapfig}
\definecolor{darkgreen}{rgb}{0,0.5,0}
\definecolor{purple}{rgb}{1,0,1}
\definecolor{Gray}{gray}{0.9}
\definecolor{lightblue}{rgb}{0.90, 0.95, 0.98}
\newcolumntype{a}{>{\columncolor{lightblue}}c}

\newcommand{\kibitz}[2]{\ifnum\Comments=1\textcolor{#1}{#2}\fi}

\usepackage[capitalize]{cleveref}
\crefname{section}{Sec.}{Secs.}
\Crefname{section}{Section}{Sections}
\Crefname{table}{Table}{Tables}
\crefname{table}{Tab.}{Tabs.}

\def\cvprPaperID{2255} 
\def\confName{CVPR}
\def\confYear{2023}

\begin{document}


\title{Supervised Masked Knowledge Distillation for Few-Shot Transformers}

\author{Han Lin\thanks{Equal contribution.$^{\hspace{1mm}\dagger}$Corresponding author.  
} , \hspace{1mm}Guangxing Han$^{*\dagger}$, \hspace{1mm}Jiawei Ma, \hspace{1mm}Shiyuan Huang, \hspace{1mm}Xudong Lin, \hspace{1mm}Shih-Fu Chang \vspace{1.5mm} \\ 
Columbia University \vspace{1mm} \\
{\tt\small \{hl3199, gh2561, jiawei.m, sh3813, xl2798, sc250\}@columbia.edu}}

\maketitle

\begin{abstract}
   Vision Transformers (ViTs) emerge to achieve impressive performance on many data-abundant computer vision tasks by capturing long-range dependencies among local features. However, under few-shot learning (FSL) settings on small datasets with only a few labeled data, ViT tends to overfit and suffers from severe performance degradation due to its absence of CNN-alike inductive bias. Previous works in FSL avoid such problem either through the help of self-supervised auxiliary losses, or through the dextile uses of label information under supervised settings. But the gap between self-supervised and supervised few-shot Transformers is still unfilled. Inspired by recent advances in self-supervised knowledge distillation and masked image modeling (MIM), we propose a novel \textbf{S}upervised \textbf{M}asked \textbf{K}nowledge \textbf{D}istillation model (SMKD) for few-shot Transformers which incorporates label information into self-distillation frameworks. Compared with previous self-supervised methods, we allow intra-class knowledge distillation on both class and patch tokens, and introduce the challenging task of masked patch tokens reconstruction across intra-class images. 
   Experimental results on four few-shot classification benchmark datasets show that our method with simple design outperforms previous methods by a large margin and achieves a new start-of-the-art. Detailed ablation studies confirm the effectiveness of each component of our model. {Code for this paper is available here: \href{https://github.com/HL-hanlin/SMKD}{https://github.com/HL-hanlin/SMKD}}.\vspace{-5mm}
\end{abstract}

\section{Introduction}




\begin{figure}[h]
    \includegraphics[width=.99\linewidth]{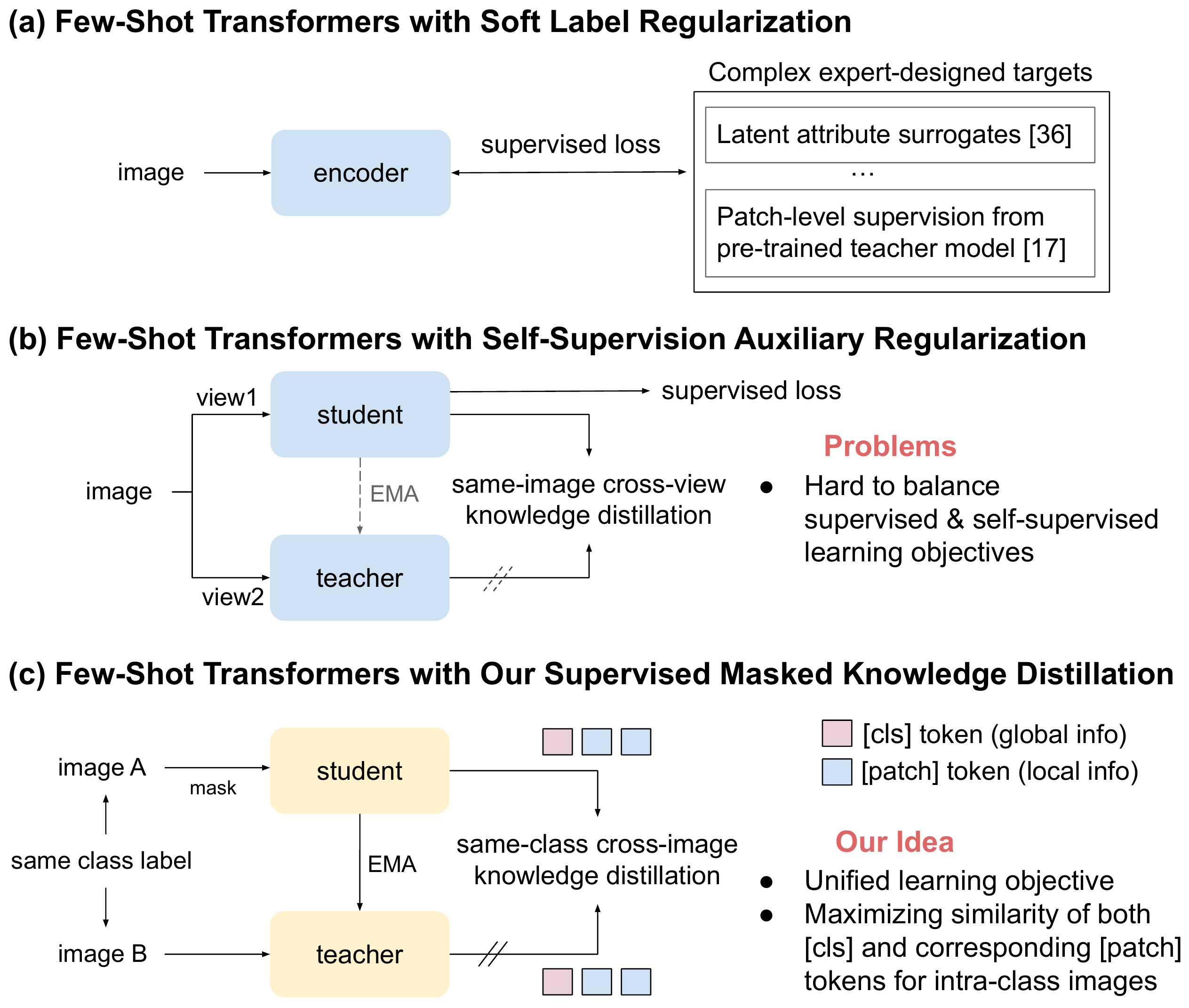}
    \vspace{-2mm}
    \caption{\small{\textbf{Comparison of the proposed idea and other existing methods for few-shot Transformers.}} Our model mitigates the overfitting issue of few-shot Transformers, by extending the masked knowledge distillation framework into the supervised setting, and enforcing the alignment of \texttt{[cls]} and corresponding \texttt{[patch]} tokens for intra-class images.}
\label{fig:front_plot}
\vspace{-4.5mm}
\end{figure}

Vision Transformers (ViTs) \cite{dosovitskiy2020image} have emerge as a competitive alternative to Convolutional Neural Networks (CNNs) \cite{he2016deep} in recent years, and have achieved impressive performance in many vision tasks including image classification \cite{dosovitskiy2020image, liu2021swin, touvron2021training, wu2021cvt}, object detection \cite{carion2020end, zhu2020deformable, dai2021up,R_RPN,han2018semi,SSD_TDR}, and object segmentation \cite{strudel2021segmenter, ranftl2021vision}. Compared with CNNs, which introduce inductive bias through convolutional kernels with fixed receptive fields \cite{krizhevsky2017imagenet}, the attention layers in ViT allow it to model global token relations to capture long-range token dependencies. However, such flexibility also comes at a cost: ViT is data-hungry and it needs to learn token dependencies purely from data. This property often makes it easy to overfit to datasets with small training set and suffer from severe generalization performance degradation \cite{liu2021efficient, lee2021vision}. 
Therefore, we are motivated to study how to make ViTs generalize well on these small datasets, especially under the few-shot learning (FSL) setting \cite{gidaris2019boosting, lu2020learning, vinyals2016matching} which aims to recognize unseen new instances at test time just from only a few (e.g. one or five) labeled samples from each new categories.

Most of the existing methods mitigate the overfitting issue of few-shot Transformers \cite{dong2022self} using various regularizations.
For instance, some works utilize label information in a weaker \cite{he2022hct}, softer \cite{ma2021partner} way, 
or use label information efficiently through patch-level supervision \cite{dong2022self}. However, these models usually design sophisticated learning targets.
On the other hand, self-distillation techniques \cite{caron2021emerging, chen2021empirical}, 
and particularly, the recent masked self-distillation \cite{he2022masked, zhou2021ibot,peng2022unified}, which distills knowledge learned from an uncorrupted image to the knowledge predicted from a masked image, have lead an emerging trend in self-supervised Transformers in various fields \cite{dong2022maskclip, xiao2022masked}. Inspired by such success,
recent works in FSL attempt to incorporate self-supervised pretext tasks into the standard supervised learning through auxiliary losses \cite{liu2021efficient, peruzzo2022spatial, mangla2020charting}, or to adopt a self-supervised pretraining, supervised training two-stage framework to train few-shot Transformers \cite{hiller2022rethinking, gani2022train}. Compared with traditional supervised methods, self-supervision can learn less biased representations towards base class, which usually leads to better generalization ability for novel classes \cite{lu2022self}. 
However, the two learning objectives of self-supervision and supervision are conflicting and it is hard to balance them during training.
Therefore, how to efficiently leverage the strengths of self-supervised learning to alleviate the overfitting issue of supervised training remains a challenge.

In this work, we propose a novel supervised masked knowledge distillation framework (SMKD) for few-shot Transformers, which handles the aforementioned challenge through a natural extension of the self-supervised masked knowledge distillation framework into the supervised setting (shown in Fig. \ref{fig:front_plot}). Different from supervised contrastive learning \cite{khosla2020supervised} which only utilizes global image features for training, we leverage multi-scale information from the images (both global \texttt{[cls]} token and local \texttt{[patch]} tokens) to formulate the learning objectives, which has been demonstrated to be effective in the recent self-supervised Transformer methods \cite{zhou2021ibot,he2022masked}.
For global \texttt{[cls]} tokens, we can simply maximize the similarity for intra-class images. However, it is non-trivial and challenging to formulate the learning objectives for local \texttt{[patch]} tokens because we do not have ground-truth patch-level annotations.
{To address this problem,} we propose to estimate the similarity between \texttt{[patch]} tokens across intra-class images using cross-attention, and enforce the alignment of corresponding \texttt{[patch]} tokens. Particularly, \textit{reconstructing masked \texttt{[patch]} tokens across intra-class images increases the difficulty of model learning, thus encouraging learning generalizable few-shot Transformer models by jointly exploiting the holistic knowledge of the image and the similarity of intra-class images.} 

\begin{figure}[t]
    \includegraphics[width=.99\linewidth]{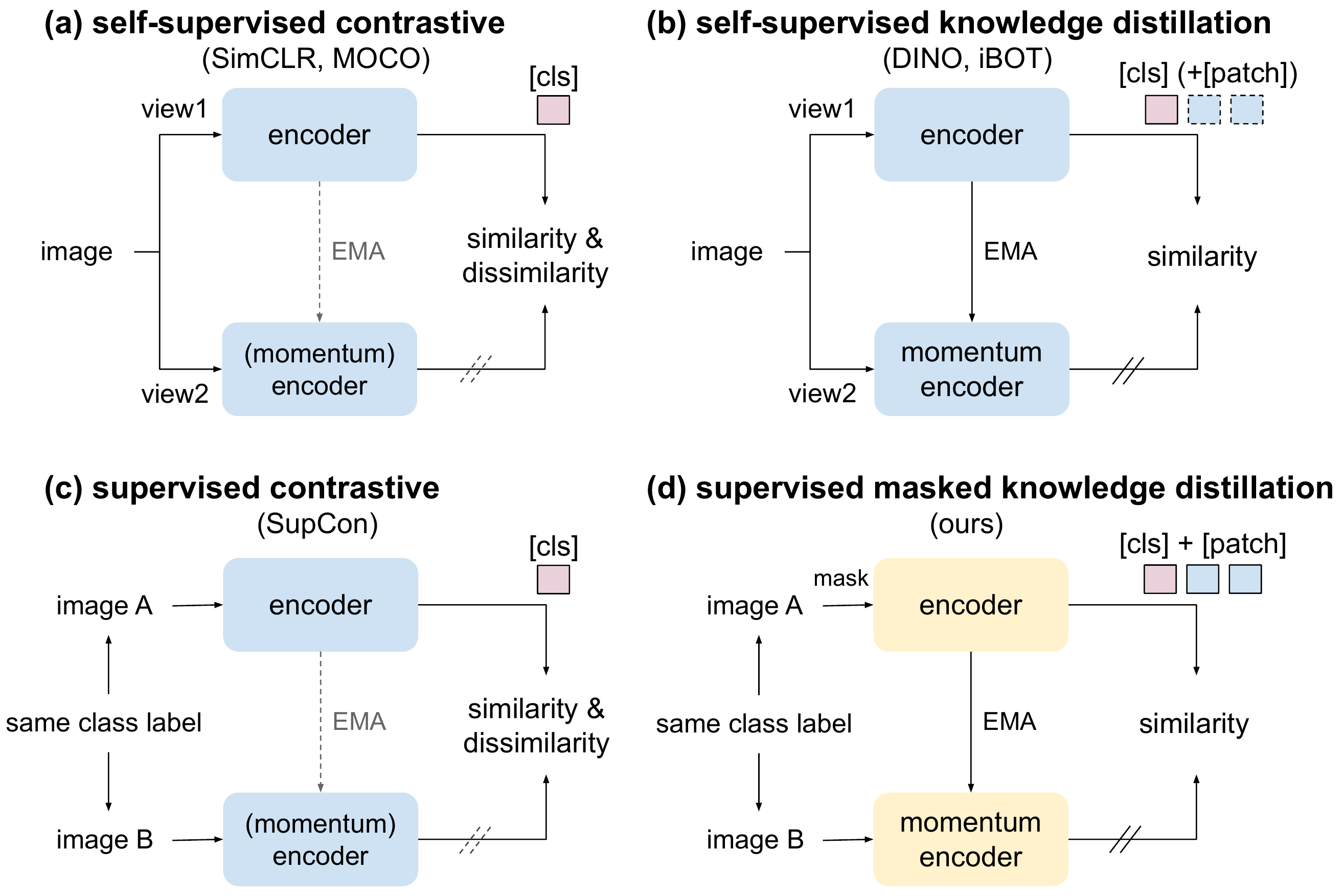}
    \vspace{-3mm}
    \caption{\small{\textbf{Comparison of other self-supervised/supervised frameworks.}  Our method (d) is a natural extension of (b) and (c), with the newly-introduced challenging task of masked \texttt{[patch]} tokens reconstruction across intra-class images. }}
\label{fig:front_plot_2}
\vspace{-2mm}
\end{figure}

As shown in Fig. \ref{fig:front_plot_2}, we compare our model with the existing self-supervised/supervised learning methods. Our model is a natural extension of the supervised contrastive learning method \cite{khosla2020supervised} and self-supervised knowledge distillation methods \cite{zhou2021ibot,caron2021emerging}. Thus our model inherits both the advantage of method \cite{khosla2020supervised} for effectively leveraging label information, and the advantages of methods \cite{zhou2021ibot,caron2021emerging} for not needing large batch size and negative samples. Meanwhile, the newly-introduced challenging task of masked \texttt{[patch]} tokens reconstruction across intra-class images makes our method more powerful for learning generalizable few-shot Transformer models.

Compared with contemporary works on few-shot Transformers \cite{he2022hct, hiller2022rethinking, dong2022self}, our framework enjoys several good properties from a practical point of view. 
(1) Our method does not introduce any additional learnable parameters besides the ViT backbone and projection head, which makes it easy to be combined with other methods \cite{he2022hct, hiller2022rethinking, Zhang_2020_CVPR}. (2) {Our method is both effective and training-efficient, with stronger performance and less training time on four few-shot classification benchmarks, compared with \cite{he2022hct, hiller2022rethinking}.}
In a nutshell, our main contributions can be summarized as follows:
\begin{itemize}
\vspace{-1mm}
    \setlength{\itemsep}{0pt}
    \setlength{\parskip}{0pt}
    \setlength{\parsep}{0pt}
    \item We propose a new supervised knowledge distillation framework (SMKD) that incorporates class label information into self-distillation, thus filling the gap between self-supervised knowledge distillation and traditional supervised learning. 
    \item Within the proposed framework, we design two supervised-contrastive losses on both class and patch levels, and introduce the challenging task of masked patch tokens reconstruction across intra-class images. 
    \item Given its simple design, we test our SMKD on four few-shot datasets, and show that it achieves a new SOTA on CIFAR-FS and FC100 by a large margin, as well as competitive performance on \emph{mini}-ImageNet and \emph{tiered}-ImageNet using the simple prototype classification method for few-shot evaluation.
\end{itemize}

\section{Related Work}

\noindent\textbf{Few-shot Learning.} Few-shot learning aims at fast knowledge transfer from seen base classes to unseen novel classes given only a few labeled samples from each novel class. The meta-learning paradigm, which simulates few-shot tasks episodically to mimic the human-learning process in the real world, once dominated FSL \cite{sung2018learning, oreshkin2018tadam, rusu2018meta, snell2017prototypical, vinyals2016matching,Han_2022_CVPR,Han_2021_ICCV,han2022meta,han2022multimodal}. Surprisingly, recent works \cite{chen2019closer, tian2020rethinking, guo2020broader,ypsilantis2021met} have shown that the state-of-the-art meta-learning methods can be outperformed by simple baseline methods using just a distance-based classifier without the complicated design of meta-learning strategies. Therefore, recent methods \cite{feng2021rethinking, he2022hct, ma2021partner} in FSL start to focus less on meta-learning, and more on learning embeddings with good generalization ability. Our paper follows this trend and proposes a knowledge distillation framework to learn generalizable embeddings.

\noindent\textbf{Vision Transformers in FSL.} Variants of Transformer \cite{vaswani2017attention} have achieved great success in NLP \cite{devlin2018bert, radford2019language}, computer vision \cite{dosovitskiy2020image, touvron2021training}, and multimodal learning \cite{radford2021learning}. 
However, the lack of inductive bias makes Transformer infamous for its data-hungry property, which is especially significant in few-shot settings when the amount of data is limited. A line of works focuses on introducing inductive bias back to the Transformer architecture, including methods using pyramid structure \cite{wang2022pvt, wang2021pyramid}, shifted windows \cite{liu2021swin}, and explicit convolutional token embeddings \cite{wu2021cvt}. With this said, some recent works \cite{hiller2022rethinking, doersch2020crosstransformers, ye2020few,Han_2022_CVPR} still show that few-shot Transformers have the potential of fast adaptation to novel classes. Our work also studies few-shot Transformers, and we show that our method can work well even on the vanilla ViT structure without explicit incorporation of inductive bias.

\noindent\textbf{Self-Supervision for FSL.} 
Self-supervised learning (SSL) has shown great potential in FSL due to its good generalization ability to novel classes. Previous methods incorporate SSL into FSL in various ways. Some works propose to include self-supervised pretext tasks into the standard supervised learning through auxiliary losses \cite{liu2021efficient, peruzzo2022spatial, mangla2020charting}. For example, \cite{liu2021efficient} proposed a regularization loss that extracts additional information from images by predicting the geometric distance between patch tokens. \cite{peruzzo2022spatial} designed their loss by discouraging spatially disordered attention maps based on the idea that objects usually occupy connected regions, and \cite{mangla2020charting} derived their auxiliary loss from self-supervision tasks of rotation and exemplar. 
Some other works \cite{hiller2022rethinking, gani2022train}, instead, adopt a two-stage procedure by pretraining a model via self-supervision before supervised training. \cite{hiller2022rethinking} takes advantage of self-supervised training with iBOT \cite{zhou2021ibot} as a pretext task, and then uses inner loop token importance reweighting for supervised fine-tuning. \cite{gani2022train} initializes its model with a pre-trained self-supervised DINO \cite{caron2021emerging}, then trained with supervised cross-entropy loss. Our work follows the second branch of work, but the difference is obvious from previous works. Instead of designing complicated training pipeline or leveraging additional learnable modules at inference time, we use supervised training on self-supervised pre-trained model, and focus mainly on bridging the gap between self-supervised and supervised knowledge distillation with minimum extra design.

Another emerging trend in SSL is Masked Image Modeling (MIM) \cite{zhou2021ibot, he2022masked, peng2022unified, bao2021beit}, which aims at recovering the patch-level target (e.g. image pixels, patch features) of the masked content in a corrupted input image. In iBOT \cite{zhou2021ibot}, the class and patch tokens share the same projection head, thus helping the ViT backbone to learn semantically meaningful patch embeddings, which can be used as a supplement to image-level global self-supervision. 

\section{Our Approach}

\begin{figure*}[t]
    \includegraphics[width=.99\linewidth]{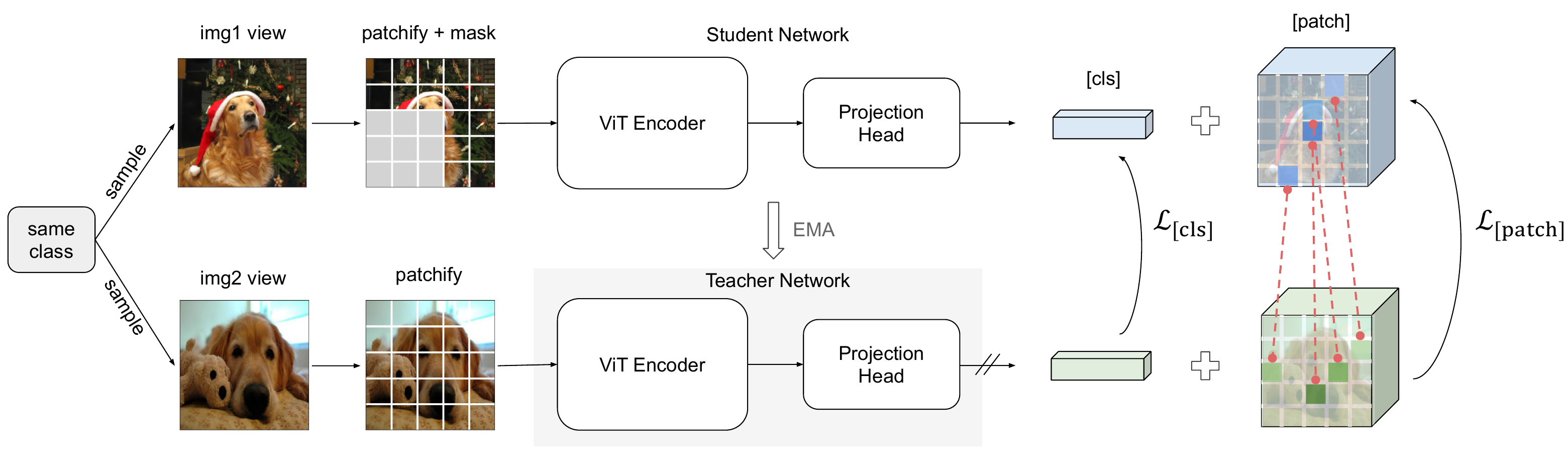}\vspace{-5mm}
    \caption{\small{\textbf{Overview of our SMKD framework}. Two views are generated from a pair of images sampled \emph{(with replacement)} from the same class. The first view, applied with random blockwise masking, is passed to the student network, while the second unmasked view is passed to the teacher network. Both networks consist of a ViT backbone and a projection head. The parameters of the teacher network are Exponentially Moving Averaged (EMA) updated by the student network. SMKD distills knowledge between intra-class cross-views on both class and patch levels. $\mathcal{L}_{[cls]}$ distills knowledge from $\texttt{[cls]}$ tokens, while $\mathcal{L}_{[patch]}$ distills knowledge from $\texttt{[patch]}$ tokens by finding dense correspondence of matched token pairs (connected by {\color{black}red} dashed lines) with highest similarities. }}
\label{fig:diagram}
\vspace{-4mm}
\end{figure*}


We present our Supervised Masked Knowledge Distillation (SMKD) framework in this section. Fig. \ref{fig:diagram} shows an overview of our model. Task definition for few-shot classification is presented in \cref{sec:task_formulation}. We explain the formulation of SSL knowledge distillation in \cref{sec:preliminary}, and how we extend it into our supervised knowledge distillation framework in \cref{sec:loss_formulation}. Training pipeline is given in \cref{sec:training_framework}.


\subsection{Learning Task Definition}\label{sec:task_formulation}

In few-shot classification, we are given two datasets $\mathcal{D}_{base}$ and $\mathcal{D}_{novel}$ with disjoint class labels $\mathcal{C}_{base}$ and $\mathcal{C}_{novel}$ ($\mathcal{C}_{base} \cap \mathcal{C}_{novel}=\varnothing$). The base dataset $\mathcal{D}_{base}$, which contains abundant labeled samples, is used to train the feature extractor. And the novel dataset $\mathcal{D}_{novel}$ is then used to sample episodes for prototype estimation and few-shot evaluation. In a standard N-way K-shot task, each episode $\mathcal{D}_{epi}=(\mathcal{D}_{S}, \mathcal{D}_{Q})$ covers N classes from $\mathcal{C}_{novel}$. The support dataset $\mathcal{D}_{S}$, which contains K samples from each class, is used for class prototype estimation, and the query dataset $\mathcal{D}_{Q}$ is then used for evaluation. This task aims at correctly classifying $\mathcal{D}_{Q}$ into N classes from sampled episodes. 
And the main focus of this paper is to train a feature extractor with good generalization ability from $\mathcal{D}_{base}$. 

\subsection{Preliminary: SSL with Knowledge Distillation}\label{sec:preliminary}

Our work is inspired by the self-supervised knowledge distillation frameworks proposed recently \cite{caron2021emerging}. Specifically, given an input image $\boldsymbol{x}$ uniformly sampled from the training set ${\mathcal{I}}$, random data augmentations are applied to generate two augmented views $\boldsymbol{x^1}$ and $\boldsymbol{x^2}$ (\emph{we represent image view indices as superscripts and patch indices as subscripts}), which are then fed into the teacher and student networks. The student network, parameterized by $\boldsymbol{\theta_s}$, consists of an encoder with ViT backbone and a projection head with a 3-layer multi-layer perceptron (MLP) followed by $l_2$-normalized bottleneck. The ViT backbone first generates a \texttt{[cls]} token, which is then entered into the projection head and outputs a probability distribution $\boldsymbol{P_{s}}^{\texttt{[cls]}}$ over K dimensions. The teacher network, parameterized by $\boldsymbol{\theta_t}$, is Exponentially Moving Averaged (EMA) updated by the student network $\boldsymbol{\theta_s}$, which distills its knowledge to the student by minimizing the cross-entropy loss over the outputs of the categorical distributions from their projection heads: 
\begin{equation}\label{eq:ss_cls}
    \mathcal{L}_{\texttt{[cls]}} = \mathcal{H}(\boldsymbol{P_{t}}^{\texttt{[cls]}}(\boldsymbol{x}^1), \boldsymbol{P_{s}}^{\texttt{[cls]}}(\boldsymbol{x}^2))
\end{equation} 
where $\mathcal{H}(x,y)=-x\log y$.

Masked Image Modeling (MIM) \cite{bao2021beit, tan2021vimpac} can be performed via self-distillation as follows \cite{zhou2021ibot}. 
Given a randomly sampled mask sequence $m\in \{0, 1\}^N$ over an image with $N$ \texttt{[patch]} tokens $\boldsymbol{x}=\{\boldsymbol{x}_i\}_{i=1}^N$, $\boldsymbol{x}_i$'s with $m_i=1$ are then replaced by a learnable token embedding $\boldsymbol{e}_{\texttt{[MASK]}}$, which results in a corrupted image $\widehat{\boldsymbol{x}}=\{\boldsymbol{\widehat{x}}_i\}_{i=1}^N=\{(1-m_i)\boldsymbol{x}_i + m_i\boldsymbol{e}_{\texttt{[MASK]}} \}_{i=1}^N$. This corrupted image and the original uncorrupted image are fed into the student and teacher networks respectively. The objective of MIM is to recover the masked tokens from the corrupted image, which is equivalent to minimizing the cross-entropy loss between the outputs of the categorical distributions of the student and teacher networks on masked patches:
\begin{equation}\label{eq:ss_mim}
    \mathcal{L}_{\text{MIM}} =  \sum_{i=1}^N m_i \cdot \mathcal{H}(\boldsymbol{P_{t}}^{\texttt{[patch]}}(\boldsymbol{x})_i, \boldsymbol{P_{s}}^{\texttt{[patch]}}(\boldsymbol{\widehat{x}})_i)
\vspace{1mm}
\end{equation}
\normalsize

\subsection{Supervised Masked Knowledge Distillation}\label{sec:loss_formulation}

\paragraph{Distill the Class Tokens.} Recent self-distillation frameworks \cite{grill2020bootstrap, caron2021emerging, zhou2021ibot} distill knowledge on \texttt{[cls]} tokens from cross-view images via Eq.(\ref{eq:ss_cls}). To incorporate label information into such self-supervised frameworks, we further allow knowledge on \texttt{[cls]} tokens to be distilled from intra-class cross-views. This can be achieved by a small extension of the way we sample images. Rather than sampling a single image $\boldsymbol{x}\sim \mathcal{I}$ and generating two views, now we sample two images $\boldsymbol{x}, \boldsymbol{y}\sim \mathcal{I}^c$ \emph{(with replacement)} and generate one view for each of them\footnote{For simplicity of illustration, we show our method with only one augmented view from each image. We sample two views from an image in our implementation. Loss is then averaged over all cross-view pairs.}. $\mathcal{I}^c\subseteq \mathcal{I}$ here denotes the set of images with the same class label $c$ in the training set $\mathcal{I}$. Specifically, we denote $\boldsymbol{{x'}}$ and $\boldsymbol{{y'}}$ as the augmented views generated from image $\boldsymbol{x}$ and $\boldsymbol{y}$ respectively. 
We apply additional random blockwise masking on $\boldsymbol{{x'}}$ and denote the resulting corrupted view as $\boldsymbol{\widehat{x}'}$. 
Then the corrupted view $\boldsymbol{\widehat{x}'}$ and uncorrupted view $\boldsymbol{{y}'}$ are sent to the student and teacher networks seperately. Now our supervised-contrastive loss on \texttt{[cls]} tokens becomes: 
\begin{equation}\label{eq:sc_cls}
    \mathcal{L}_{\texttt{[cls]}} =  \mathcal{H}(\boldsymbol{P_{t}}^{\texttt{[cls]}}(\boldsymbol{y'}),\boldsymbol{P_{s}}^{\texttt{[cls]}}(\boldsymbol{\widehat{x}}'))
\end{equation}

When $\boldsymbol{x}, \boldsymbol{y}$ are sampled as the same image ($\boldsymbol{x}=\boldsymbol{y}$), our loss performs \emph{masked} self-distillation across two views of the same image similar to Eq.(\ref{eq:ss_cls}). In the other situation when $\boldsymbol{x}$ and $\boldsymbol{y}$ represent different images ($\boldsymbol{x}\neq \boldsymbol{y}$), our loss then minimizes the cross-entropy loss of projected \texttt{[cls]} tokens among all cross-view pairs between images $\boldsymbol{x}$ and $\boldsymbol{y}$.

Such design has two major advantages. (1) It can be implemented efficiently. Instead of intentionally sampling image pairs from the same class, we just need to look through the images in a mini-batch, find image pairs belonging to the same class, and then apply our loss in Eq.(\ref{eq:sc_cls}). (2) 
Unlike previous works that use either supervised \cite{ma2021partner, zhao2020makes} or self-supervised \cite{chen2020simple, he2020momentum} contrastive losses, our method follows the recent trend in SSL works \cite{zbontar2021barlow, chen2021exploring, grill2020bootstrap, caron2021emerging} and avoids the need for negative examples.\\

\noindent\textbf{Distill the Patch Tokens.} 
Beyond the knowledge distillation on the global \texttt{[cls]} tokens, we introduce the challenging task of masked patch tokens reconstruction across intra-class images, to fully exploit the local details of the image for training.
Our main intuition here is based on the following hypothesis: for intra-class images, even though their semantic information can be vastly different at the patch level, there should at least exist some patches that share similar semantic meanings. 

For each patch $k$ from an input view $\boldsymbol{y'}$ sent to the teacher network (with its corresponding token embedding defined as $\boldsymbol{f}^{\boldsymbol{t}}_{k}$), we need to first find the most similar patch $k^+$ from the masked view $\boldsymbol{\widehat{x}'}$ of the student network (with its corresponding token embedding defined as $\boldsymbol{f}^{\boldsymbol{s}}_{k^+}$), and then perform knowledge distillation between these two matched tokens. The token embedding $\boldsymbol{f}^{\boldsymbol{s}}_{k^+}$ represents either the ${k^+}^{th}$ \texttt{[patch]} token if patch $k^+$ is unmasked, or the reconstructed patch token if it is masked. 

As we do not have any patch-level annotations, we use cosine similarity to find the best-matched patch of $k$ among all the \texttt{[patch]} tokens in the student network: \vspace{-1mm}
\begin{equation}
    k^+ = \arg \max_{l\in[N]}  \frac{{\boldsymbol{f}^{\boldsymbol{t}}_k}^{\top}\boldsymbol{f}^{\boldsymbol{s}}_l}{ \|{\boldsymbol{f}^{\boldsymbol{t}}_k}\|\|\boldsymbol{f}^{\boldsymbol{s}}_l\|}
\end{equation}
Our patch-level knowledge distillation loss now becomes: \vspace{-1mm}
\small
\begin{align}\label{eq:sc_patch}
\mathcal{L}_{\texttt{[patch]}} =  
       \sum_{k=1}^N  {\omega_{k^+}}\cdot \mathcal{H}(\boldsymbol{P_{t}}^{\texttt{[patch]}}(\boldsymbol{{y}'})_k, \boldsymbol{P_{s}}^{\texttt{[patch]}}(\boldsymbol{\widehat{x}'})_{k^+})
\end{align}
\normalsize
\noindent{where $\omega_{k^+}$ is a scalar representing the weight we give to each loss term. We find that taking it as a constant value is effective enough. The ablation study of more complex designs of $\omega_{k^+}$ is given in the appendix.}

Our loss shares some similarities with DenseCL \cite{wang2021dense}. However, the differences are also obvious: (1) Our loss serves as an extension of their self-supervised contrastive loss into a supervised variant.
(2) We further incorporate MIM into our design and allow masked patches to be matched, which makes our task harder and leads to more semantically meaningful patch embeddings.

\subsection{Training Pipeline}\label{sec:training_framework}
We train our model in two stages: self-supervised pretraining and supervised training. 
In the first stage, we use the recently proposed MIM framework \cite{zhou2021ibot} for self-supervised pretraining. The self-supervised loss is the summation of $\mathcal{L}_{\texttt{[cls]}}$ and $\mathcal{L}_{\text{MIM}}$ in Eq.(\ref{eq:ss_cls}) and Eq.(\ref{eq:ss_mim}) without scaling. 
In the second stage, we continue training the pre-trained model using our supervised-contrastive losses $\mathcal{L}_{\texttt{[cls]}}$ and $\mathcal{L}_{\texttt{[patch]}}$ from Eq.(\ref{eq:sc_cls}) and Eq.(\ref{eq:sc_patch}). We define our training loss as: $\mathcal{L}=\mathcal{L}_{\texttt{[cls]}}+\lambda \mathcal{L}_{\texttt{[patch]}}$, where $\lambda$ controls the relative scale of these two components. A relatively large $\lambda$ will make our model focus more on localization and less on high-level semantics.

\section{Experiment}

\subsection{Datasets}

Our model is evaluated on four widely used and publicly available few-shot classification datasets: \emph{mini}-ImageNet \cite{vinyals2016matching}, \emph{tiered}-ImageNet \cite{ren2018meta}, CIFAR-FS \cite{bertinetto2018meta}, and FC100 \cite{oreshkin2018tadam}. \emph{mini}-ImageNet and \emph{tiered}-ImageNet are derived from ImageNet \cite{deng2009imagenet}, while CIFAR-FS and FC100 are derived from CIFAR100 \cite{krizhevsky2009learning}. \emph{mini}-ImageNet contains 100 classes, which are randomly split into 64 base classes for training, 16 classes for few-shot validation, and 20 classes for few-shot evaluation. There are 600 images in each class. \emph{tiered}-ImageNet contains 609 classes with 779165 images in total. The class split for training, few-shot validation and few-shot evaluation are 351, 97, and 160. CIFAR-FS contains 100 classes with class split as 64, 16, and 20. FC100 contains 100 classes with class split as 60, 20, and 20. For both of these two datasets, each class has 600 images with smaller resolutions (32 $\times$ 32) compared with ImageNet.

\subsection{Implementation Details}
\textbf{Self-supervised pretraining.} We pre-train our Vision Transformer backbone and projection head following the same pipeline in iBOT \cite{zhou2021ibot}. Most of the hyper-parameter settings are kept unchanged without tuning. We use a batch size of 640 and a learning rate of 0.0005 decayed with the cosine schedule. \emph{mini}-ImageNet and \emph{tiered}-ImageNet are pre-trained for 1200 epochs, and CIFAR-FS and FC100 are pre-trained for 900 epochs. All models are trained on 8 Nvidia RTX 3090 GPUs. Detailed training parameter settings are included in the appendix. 

\textbf{Supervised knowledge distillation} After getting the pre-trained model, we train it with our supervised-contrastive losses. 
We find that the model with the best performance can converge within 60 epochs in the validation set. We use the same batch size and learning rate as in the pretraining stage. Compared with the first stage, the only extra hyper-parameter is the scaling parameter $\lambda$. We set it to make the ratio between $\mathbf{\mathcal{L}_{\texttt{[patch]}}}$ and $\mathbf{\mathcal{L}_{\texttt{[cls]}}}$ roughly around 2. Ablation over different $\lambda$ is given in the appendix. 

\begin{table*}[h!]
\begin{center}
\setlength{\tabcolsep}{0.7em}
\renewcommand{\arraystretch}{1.0}
\caption{\small{\textbf{Results on mini-ImageNet and tiered-ImageNet.} Top three methods are colored in {\color{red} red}, {\color{blue} blue} and {\color{green} green} respectively. A more comprehensive version of this table is shown in the appendix.}}
\label{table:results_imagenet}
\vspace{-3mm}
\scalebox{0.97}{
\begin{tabular}{ccccc|cc}
\hlineB{3}
\multirow{2}{*}{Method} & \multirow{2}{*}{Backbone} & \multirow{2}{*}{\#Params} & \multicolumn{2}{c|}{miniImageNet,5-way} & \multicolumn{2}{c}{tieredImageNet,5-way} \\ \cline{4-7} 
 & & & 1-shot & 5-shot & 1-shot & 5-shot  \\ \hline
 DeepEMD~\cite{Zhang_2020_CVPR} & \emph{ResNet-12} & 12.4M & ${{65.91\pm0.82}}$  & ${{82.41\pm0.56}}$ & ${{71.16\pm0.87}}$ & ${{86.03\pm0.58}}$ \\
 IE~\cite{Rizve_2021_CVPR} & \emph{ResNet-12} & 12.4M &  ${{67.28\pm0.80}}$  & ${{84.78\pm0.52}}$ & ${{72.21\pm0.90}}$ & ${{87.08\pm0.58}}$ \\
 COSOC~\cite{luo2021rectifying} & \emph{ResNet-12} & 12.4M & ${{69.28\pm0.49}}$  & ${{85.16\pm0.42}}$ & ${{73.57\pm0.43}}$ & ${{87.57\pm0.10}}$ \\
 Meta-QDA~\cite{zhang2021shallow} & \emph{WRN-28-10} & 36.5M & ${{67.38\pm0.55}}$  & ${{84.27\pm0.75}}$ & ${{74.29\pm0.66}}$ & ${{89.41\pm0.77}}$ \\
 OM~\cite{qi2021transductive} & \emph{WRN-28-10} & 36.5M & ${{66.78\pm0.30}}$  & ${{85.29\pm0.41}}$ & ${{71.54\pm0.29}}$ & ${{87.79\pm0.46}}$ \\
 \hline
 SUN~\cite{dong2022self} & \emph{ViT} & 12.5M & ${{67.80\pm0.45}}$  & ${{83.25\pm0.30}}$ & ${{72.99\pm0.50}}$ & ${{86.74\pm0.33}}$ \\  
 FewTURE~\cite{hiller2022rethinking} & \emph{Swin-Tiny} & 29.0M & ${{72.40\pm0.78}}$  & ${{86.38\pm0.49}}$ & ${{76.32\pm0.87}}$ & ${{89.96\pm0.55}}$ \\ 
 HCTransformers~\cite{he2022hct} & 3$\times$\emph{ViT-S} & 63.0M & ${\color{blue}{74.74\pm0.17}}$ & ${\color{blue}{89.19\pm0.13}}$ & ${\color{blue}{79.67\pm0.20}}$ & ${\color{red}{91.72\pm0.11}}$ \\  
 \hline
Ours (Prototype)  & \emph{ViT-S} & 21M & ${\color{green}{74.28\pm0.18}}$  & ${{88.82\pm0.09}}$  &    ${\color{green}{78.83\pm0.20}}$     &  ${{91.02\pm0.12}}$       \\
Ours (Classifier)  & \emph{ViT-S} & 21M & ${{74.10\pm0.17}}$  & ${\color{green}{88.89\pm0.09}}$  &    ${{78.81\pm0.21}}$     &  ${\color{green}{91.21\pm0.11}}$       \\ 
Ours + HCT~\cite{he2022hct}  & 3$\times$\emph{ViT-S} & 63M & ${\color{red}{75.32\pm0.18}}$  & ${\color{red}{89.57\pm0.09}}$  &    ${\color{red}{79.74\pm0.20}}$     &  ${\color{blue}{91.68\pm0.11}}$       \\ 
\hlineB{3}
\end{tabular}}
\end{center}
\vspace{-4mm}
\end{table*}
\begin{table*}[h]
\begin{center}
\setlength{\tabcolsep}{0.7em}
\renewcommand{\arraystretch}{1.0}
\caption{\small{\textbf{Results on CIFAR-FS and FC100}.} A more comprehensive version of this table is shown in the appendix.}
\label{table:results_cifar}
\vspace{-3mm}
\scalebox{0.97}{
\begin{tabular}{ccccc|cc}
\hlineB{3}
\multirow{2}{*}{Method} & \multirow{2}{*}{Backbone} & \multirow{2}{*}{\#Params} & \multicolumn{2}{c|}{CIFAR-FS,5-way} & \multicolumn{2}{c}{FC100,5-way} \\ \cline{4-7} 
 & & & 1-shot & 5-shot & 1-shot & 5-shot  \\ \hline
BML~\cite{zhou2021binocular} & \emph{ResNet-12} & 12.4M & ${{73.45\pm0.47}}$  & ${{88.04\pm0.33}}$ & ${{45.00\pm0.41}}$  & ${{63.03\pm0.41}}$ \\
IE~\cite{Rizve_2021_CVPR} & \emph{ResNet-12} & 12.4M & ${{77.87\pm0.85}}$  & ${{89.74\pm0.57}}$ & ${{47.76\pm0.77}}$  & ${{65.30\pm0.76}}$ \\
TPMN~\cite{wu2021task} & \emph{ResNet-12} & 12.4M  & ${{75.50\pm0.90}}$  & ${{87.20\pm0.60}}$ & ${{46.93\pm0.71}}$  & ${{63.26\pm0.74}}$ \\
PSST~\cite{chen2021pareto} & \emph{WRN-28-10} & 36.5M & ${{77.02\pm0.38}}$  & ${{88.45\pm0.35}}$ & - & -  \\
Meta-QDA~\cite{zhang2021shallow} & \emph{WRN-28-10} & 36.5M &  ${{75.95\pm0.59}}$  & ${{88.72\pm0.79}}$ & - & -  \\
\hline
SUN~\cite{dong2022self} & \emph{ViT} & 12.5M & ${{78.37\pm0.46}}$  & ${{88.84\pm0.32}}$ & - & -  \\   
FewTURE~\cite{hiller2022rethinking} & \emph{Swin-Tiny} & 29.0M & ${{77.76\pm0.81}}$  & ${{88.90\pm0.59}}$ & ${{47.68\pm0.78}}$  & ${{63.81\pm0.75}}$  \\  
HCTransformers~\cite{he2022hct} & 3$\times$\emph{ViT-S} & 63.0M & ${\color{green}{78.89\pm0.18}}$ & ${\color{green}{90.50\pm0.09}}$ & ${\color{green}{48.27\pm0.15}}$ & ${\color{green}{66.42\pm0.16}}$ \\  
\hline
Ours (Prototype) & \emph{ViT-S} & 21M & ${\color{red}{80.08\pm0.18}}$   &  ${\color{blue}{90.63\pm0.13}}$   & ${\color{red}{50.38\pm0.16}}$   &  ${\color{blue}{68.37\pm0.16}}$  \\
Ours (Classifier)  & \emph{ViT-S} & 21M & ${\color{blue}{79.82\pm0.18}}$  & ${\color{red}{90.91\pm0.13}}$  &    ${\color{blue}{50.28\pm0.16}}$     &  ${\color{red}{68.50\pm0.16}}$       \\ 
\hlineB{3}
\end{tabular}}
\end{center}
\vspace{-6mm}
\end{table*}

\textbf{Few-shot Evaluation} We use the simple prototype classification method (\emph{Prototype}) \cite{snell2017prototypical, oreshkin2018tadam} and the linear classifier (\emph{Classifier}) used in S2M2 \cite{mangla2020charting} as our default methods for few-shot evaluation. For each sampled episode data in an N-way K-shot task, \emph{Prototype} first estimates each class prototype using the averaged feature over K support samples. Then a new sampled query image is classified into one of the N classes which has the highest cosine similarity between its feature vector and the class prototype. Instead, \emph{Classifier} trains a linear classifier from the N$\times$K support samples, which is then used to classify new query samples. More complex evaluation methods (e.g. DeepEMD) are also compatible with our framework. The feature we used for evaluation is the concatenation of the \texttt{[cls]} token with the weighted average \texttt{[patch]} token (\emph{weighted avg pool}). The weights of \emph{weighted avg pool} is the average of the self-attention values of the \texttt{[cls]} token with all heads of the last attention layer. Ablation over different choices of features for evaluation is studied in \cref{sec:ablation_tokens}.

\subsection{\hspace{-0.4mm}Comparison With the State-of-the-arts (SOTAs)}
We evaluate our proposed SMKD with the above-mentioned evaluation methods on four few-shot classification datasets. Our goal in this section is to prove the effectiveness of our method given its simple training pipeline and evaluation procedure. This being said, the performance of our method can be further boosted by adopting strategies from contemporary works (see the last row "Ours+HCT\cite{he2022hct}" in Table \ref{table:results_imagenet} and detailed results in Table \ref{table:our_best}).



Compared with traditional convolutional backbones, Transformer-based models are still underdeveloped in few-shot classification. Recent methods with Transformer backbones \cite{hiller2022rethinking, he2022hct, dong2022self} differ a lot in their training pipelines and evaluation procedures. SUN \cite{dong2022self} first pre-trains a teacher network with supervised loss, then uses it to generate patch-level supervision for the student network as a supplement to class-level supervision. HCTransformers \cite{he2022hct} converts class-level prediction into a latent prototype learning problem, and introduces spectral tokens pooling to merge neighboring tokens with similar semantic meanings adaptively. FewTURE \cite{hiller2022rethinking}, which adopts a similar two-stage training pipeline as ours, shows the effectiveness of inner loop token importance reweighting to avoid supervision collapse. 

Despite the success of these works, our SMKD still shows competitive performance given its simple design. Table \ref{table:results_imagenet} contains the results of \emph{mini}-ImageNet and \emph{tiered}-ImageNet. With \emph{Prototype} and \emph{Classifier} as the evaluation method, our proposed SMKD outperforms all methods with \emph{ResNet} and \emph{WRN} backbones, and ranked second among methods with Transformer backbones. By adopting the same strategies (patch size of 8 and spectral tokens pooling) as in HCTransformers\cite{he2022hct}, our method achieves a new SOTA on \emph{mini}-ImageNet. The effect of each of these two strategies is shown in Table \ref{table:our_best}. Furthermore, as displayed in Table \ref{table:results_cifar}, our method performs the best on the two small-resolution datasets (CIFAR-FS and FC100), and outperforms all previous results by great margins (\textbf{0.93\%} for 1-shot and \textbf{0.41\%} for 5-shot on CIFAR-FS, and \textbf{over 2\%} for both 1 and 5-shot on FC100). \emph{In conclusion, our method grows more effective on datasets with smaller resolutions and fewer training images.}

\begin{table}[h!]
    \begin{center}
    \caption{\small{\textbf{Results for small patch size and spectral tokens pooling}. Results are evaluated on \emph{mini}-ImageNet with \emph{Prototype} and \emph{Classifier} methods, and the best is reported.}\vspace{-3mm}}
    \label{table:our_best}
    \scalebox{0.90}{
     \begin{tabular}{c@{\hskip -0.002in}cc@{\hskip -0.005in}cc} 
    \toprule
    Method &  Patch 8 & Spec. Pool  & 1-shot & 5-shot  \\  
    \toprule
     HCT~\cite{he2022hct} & \checkmark &  & ${{71.27\pm0.17}}$ & ${{84.68\pm0.10}}$  \\
     HCT~\cite{he2022hct} & \checkmark & \checkmark & ${{74.62\pm0.20}}$ & ${{89.19\pm0.13}}$  \\
     \hline
     Ours &  &  & ${{74.28\pm0.18}}$  & ${{88.82\pm0.09}}$ \\ 
     Ours & \checkmark & & ${{74.00\pm0.17}}$ & ${\mathbf{89.57\pm0.09}}$ \\ 
     Ours & \checkmark & \checkmark & $\mathbf{75.32\pm0.18}$ & $89.23\pm0.18$ \\ 
    \bottomrule
    \end{tabular}}
    \end{center}
\vspace{-6mm}
\end{table}
\normalsize


The top competitor to our model is HCTransformer \cite{he2022hct}. However, their method is more complex compared with ours in the following two aspects: (1) Their method uses a smaller patch size of 8 rather than 16 (which is used in most other methods as well as ours). As shown in Table \ref{table:comparison_training_clock_time}, this makes their training procedure 4$\times$ slower than ours in each epoch. (2) Their method contains three sets of cascaded transformers, each corresponding to a set of teacher-student network with ViT-S backbone. This triples the total number of learnable parameters (63M versus 21M in ours) and incurs much longer inference time.

\begin{table}[h!]
    \begin{center}
    \caption{\small{\textbf{Training clock time comparison on \emph{mini}-ImageNet}. For HCT, we report the result provided in their paper. All experiments are conducted on 8 Nvidia RTX 3090 GPUs. }\vspace{-3mm}}
    \label{table:comparison_training_clock_time}
    \scalebox{0.82}{
     \begin{tabular}{c@{\hskip 0.05in}cccccc} 
    \toprule
     & \multicolumn{2}{c}{Stage1}  & \multicolumn{2}{c}{Stage2} & \multicolumn{2}{c}{Stage3} \\
    \cline{2-7} 
    HCT~\cite{he2022hct} & \multicolumn{2}{c}{21.1h (400 epoch)} & \multicolumn{2}{c}{0.58h (2 epoch) } & \multicolumn{2}{c}{0.21h (2 epoch)} \\
    \bottomrule
     &  \multicolumn{3}{c}{Self-Supervised Pre-train}  & \multicolumn{3}{c}{Supervised Training} \\
    \cline{2-7} 
    Ours & \multicolumn{3}{c}{15.0h (1200 epoch)} & \multicolumn{3}{c}{1.08h (60 epoch)} \\
    \bottomrule
    \end{tabular}}
    \end{center}
\vspace{-4mm}
\end{table}
\normalsize

For a fair comparison, we compare few-shot classification results of methods with the same ViT-S backbone (21M parameters) on \emph{mini}-ImageNet in Table \ref{table:comparison_same_backbone}. The results for HCT\cite{he2022hct} and FewTURE\cite{hiller2022rethinking} are copy-pasted from their paper. Our model outperforms them by a large margin when the number of trainable parameters is comparable. 

\begin{figure}[h]
    \includegraphics[width=.99\linewidth]{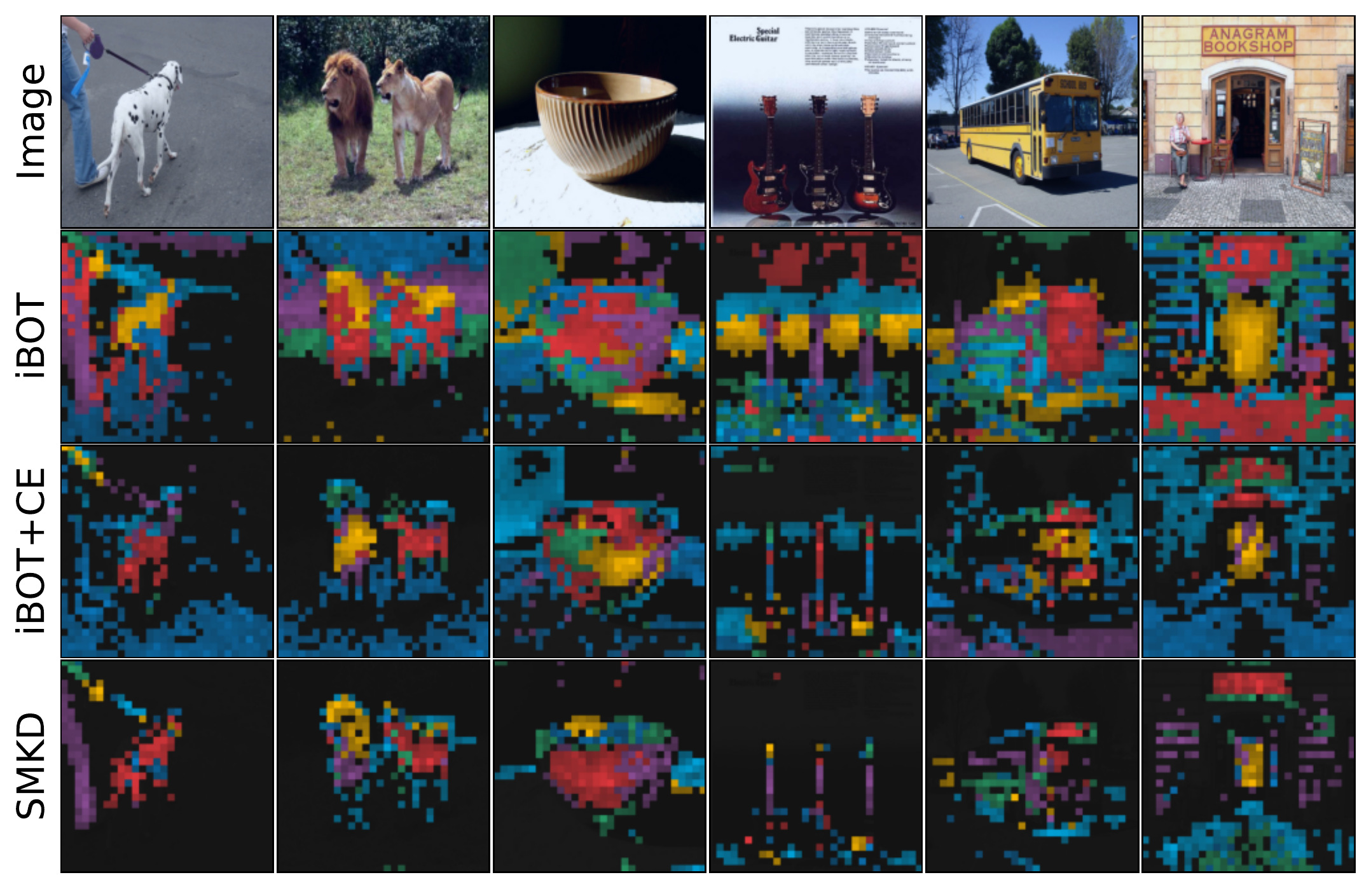}
    \vspace{-3mm}
    \caption{\small{\textbf{Visualization of multi-head self-attention maps}. The self-attention of the $\texttt{[cls]}$ tokens with different heads in the last attention layer of ViT are visualized in different colors. iBOT+CE represents the model first pre-trained with iBOT, then trained with CE loss. Our SMKD pays more attention to the foreground objects, especially the most discriminative parts.}}
\label{fig:attention}
\vspace{-2mm}
\end{figure}
\begin{figure}[h]
    \includegraphics[width=.99\linewidth]{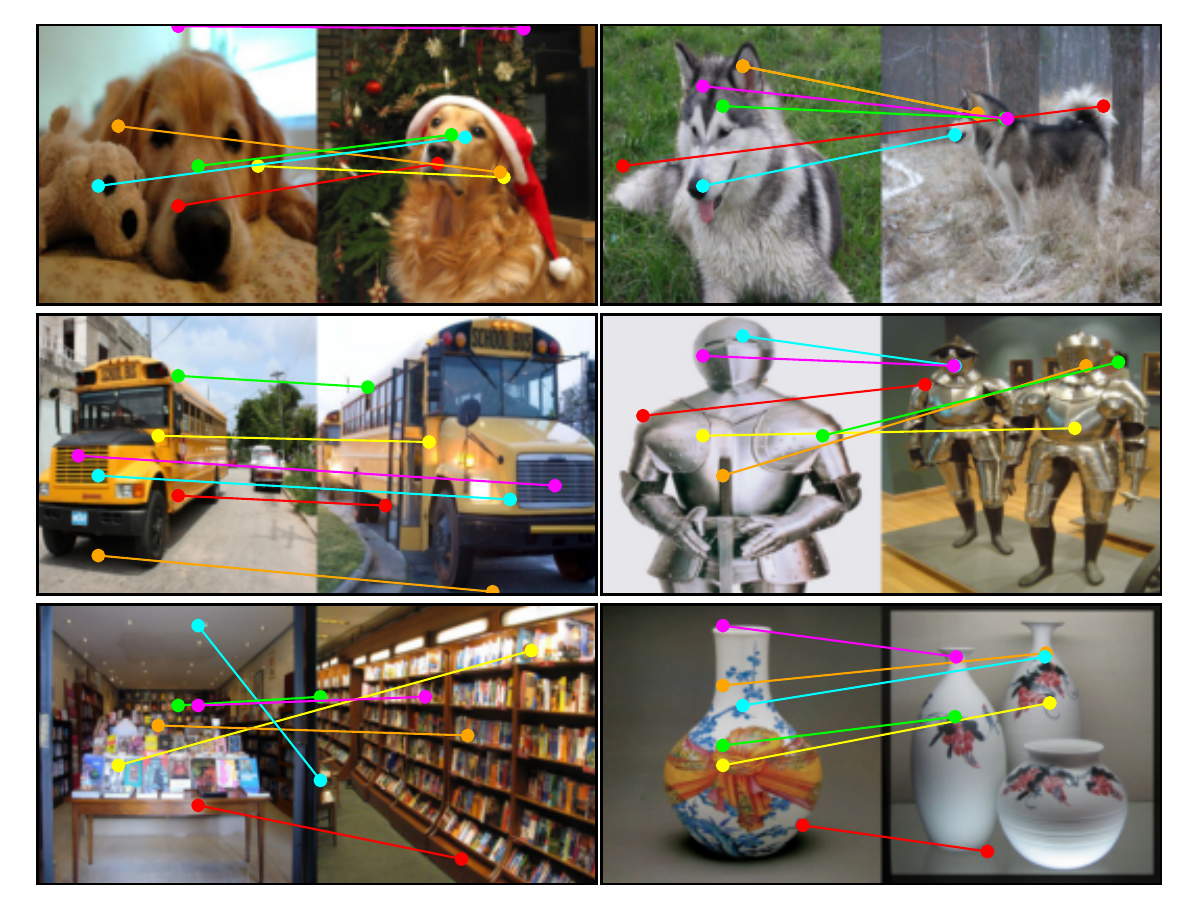}
    \vspace{-4mm}
    \caption{\small{\textbf{Visualization of dense correspondence}. We use the patches with the highest self-attention of the $\texttt{[cls]}$ token on each attention head (6 in total) of the last layer of ViT-S as queries. Best matched patches with highest similarities are connected with lines. }}
\label{fig:dense_match}
\vspace{-4mm}
\end{figure}

Visualizations of the multi-head self-attention and patch token correspondence are shown in Fig. \ref{fig:attention} and Fig. \ref{fig:dense_match}.

\begin{table}[h!]
    \begin{center}
    \caption{\small{\textbf{Comparison results with the same ViT-S backbone on \emph{mini}-ImageNet}. For HCT, we report their result of the first student transformer (training time shown in Stage 1 of Table \ref{table:comparison_training_clock_time}). }\vspace{-3mm}}
    \label{table:comparison_same_backbone}
    \scalebox{0.9}{
     \begin{tabular}{ccccc} 
    \toprule
    Method &  Backbone  & 1-shot & 5-shot  \\ 
    \toprule
     HCT~\cite{he2022hct} & ViT-S & $71.27\pm0.17$ & $84.68\pm0.10$ \\
     FewTURE~\cite{hiller2022rethinking} & ViT-S  & $68.02\pm0.88$ & $84.51\pm0.53$ \\ 
    \bottomrule
    \rowcolor{lightblue}
    ${\textrm{Ours}}$ &  ViT-S  & ${{74.28\pm0.18}}$  & ${{88.82\pm0.09}}$    \\
    \bottomrule
    \end{tabular}}
    \end{center}
\vspace{-4mm}
\end{table}
\normalsize

\subsection{Ablation Study}\label{sec:ablation_study}

\noindent\textbf{Self-Supervised MIM Pretraining Helps.}\label{sec:ss_is_necessary} Our training pipeline contains two stages. A natural question is whether the first self-supervised pretraining stage is necessary. We can see from Table \ref{table:ablation_pretraining_stage} that the classification accuracy of our method drops a lot without pretraining. If we train in a single stage, with the combination of the self-supervised learning losses and supervised contrastive learning losses, the model is still hard to converge without a good initialization (1-shot: 43.71 vs. our 74.28.
5-shot: 60.27 vs. our 88.82 on \emph{mini}-ImageNet). An explanation is that it will be hard for our patch-level knowledge distillation to find matched patches if the ViT backbone is initialized with random weights. What we observe is that our model finds some shortcut solution that prohibits it from proper training. Additionally, we also test the traditional CE loss with and without self-supervised pretraining. The model pre-trained with self-supervision and then trained with supervised CE loss works surprisingly well and even outperforms most of the methods in Table \ref{table:results_imagenet} on \emph{mini}-ImageNet. This further supports the usefulness of our two-stage training procedure. For the masking strategy in the first pretraining stage, our ablation study over three masking types (blockwise, random, and no mask) in the appendix shows that blockwise masking balances the best between 1 and 5-shots classification accuracy.

\begin{table}[h!]
    \begin{center}
    \caption{\small{\hspace{-1mm}\textbf{Few-shot evaluation w/wo self-supervised pretraining.} For models with pretraining stage, we pre-train with iBOT for 1200 epochs, then train with CE or our loss for 60 epochs. For models without pretraining stage, we train with supervised losses for 60 epochs directly. Results are evaluated by \emph{Prototype} classification method on \emph{mini}-ImageNet.}}
    \label{table:ablation_pretraining_stage}
    \vspace{-3mm}
    \scalebox{0.95}{
     \begin{tabular}{cccc} 
    \toprule
     Method & Pre-train & 1-shot  & 5-shot \\ 
    \toprule
     CE & & ${{48.98\pm0.16}}$ & ${{67.62\pm0.14}}$  \\
     \rowcolor{lightblue}
     CE & \checkmark & ${{71.02\pm0.18}}$ & ${{87.25\pm0.10}}$  \\
     \hline 
     Ours & & ${{33.14\pm0.13}}$ & ${{45.02\pm0.14}}$  \\
     \rowcolor{lightblue}
     Ours & \checkmark & ${{74.28\pm0.18}}$  & ${{88.82\pm0.09}}$  \\
    \bottomrule
    \end{tabular}}
    \end{center}
\vspace{-5mm}
\end{table}
\normalsize

\begin{table}[h!]
    \begin{center}
    \caption{\small{\textbf{Ablation of different losses in the supervised training stage.} All losses are trained for 60 epochs initialized with the pre-trained iBOT model. The first row represents the pre-trained iBOT without additional supervised training. ${{\mathcal{L}_{CE}}+{\mathcal{L}_{\texttt{[patch]}}}}$ stands for the combination of CE loss and our patch-level loss. Results are evaluated by \emph{Prototype} method on \emph{mini}-ImageNet.}}
    \label{table:ablation_phase2_loss}
    \vspace{-3mm}
    \scalebox{0.95}{
     \begin{tabular}{ c c c} 
    \toprule
     Loss & 1-shot  & 5-shot \\ 
    \toprule
     - & ${{60.93\pm0.17}}$ & ${{80.38\pm0.12}}$  \\
     ${{\mathcal{L}_{CE}}}$ &  ${{71.02\pm0.18}}$ & ${{87.25\pm0.10}}$  \\
     ${{\mathcal{L}_{\texttt{[cls]}}}}$ &  ${{70.21\pm0.17}}$ & ${{87.03\pm0.10}}$  \\
     ${{\mathcal{L}_{\texttt{[patch]}}}}$ & ${{70.84\pm0.18}}$ & ${{85.90\pm0.11}}$  \\
     ${{\mathcal{L}_{CE}}+{\mathcal{L}_{\texttt{[patch]}}}}$ & ${{70.70\pm0.18}}$  & ${{86.77\pm0.10}}$  \\
     \rowcolor{lightblue}
     ${{\mathcal{L}_{[\texttt{cls}]}}+{\mathcal{L}_{\texttt{[patch]}}}}$ & ${{74.28\pm0.18}}$  & ${{88.82\pm0.09}}$  \\
    \bottomrule
    \end{tabular}}
    \end{center}
\vspace{-6mm}
\end{table}
\normalsize

\noindent\textbf{Class and Patch-Level Distillation Complement Each Other.} \label{sec:ablation_sc_objectives}
Our SMKD uses the combination of class-level and patch-level supervised-contrastive losses as the final objective. Here we study several different combinations of supervised losses to confirm our proposed method performs the best. Results in Table \ref{table:ablation_phase2_loss} show that combining $\mathcal{L}_{\texttt{[cls]}}$ with $\mathcal{L}_{\texttt{[patch]}}$ achieves \textbf{13.35\%} improvement over baseline for 1-shot and \textbf{8.44\%} for 5-shot classification, which outperforms each of its components. Sharing the parameters of projection heads for class and patch tokens allows the semantics obtained from their distillation to complement each other, which leads to much better performance. This can also be seen from the comparison between $\mathcal{L}_{CE}+\mathcal{L}_{\texttt{[patch]}}$ and $\mathcal{L}_{\texttt{[cls]}}+\mathcal{L}_{\texttt{[patch]}}$. $\mathcal{L}_{CE}$ alone achieves stronger performance than $\mathcal{L}_{\texttt{[cls]}}$, but combining it with $\mathcal{L}_{\texttt{[patch]}}$ does not provide additional benefits with separate classification heads. \\

\noindent\textbf{Weighted Average Pooling Boosts Performance.} \label{sec:ablation_tokens}
Some recent works \cite{xu2021videoclip, touvron2022deit} find that using average pooling instead of the \texttt{[cls]} token during evaluation can encourage token-level tasks (e.g. localization, segmentation). Inspired by such findings, we evaluate our model with different tokens as well as their combinations. Results from Table \ref{table:ablation_tokens} show that the performance of the \texttt{[cls]} token can be boosted by \textbf{3.02\%} for 1-shot and \textbf{0.79\%} for 5-shot by just concatenating with the weighted average pooling token. This shows that our proposed method can learn meaningful representations for both \texttt{[cls]} and \texttt{[patch]} tokens. Moreover, weighted average pooling and average pooling tokens share similar information, but the former performs better because it gives less weight to the background.

\vspace{-1mm}
\begin{table}[h!]
    \begin{center}
    \caption{\small{\textbf{Ablation of different tokens for few-shot evaluation.} "cls" stands for the \texttt{[cls]} token, "avg pool" stands for using average pooling as token, "weighted avg pool" stands for weighted average \texttt{[patch]} token with weights from the self-attention module from the last block of ViT. The last four rows represent concatenation of these three tokens. All tokens are normalized to unit length. Results are evaluated by \emph{Prototype} classification method.}}
    \label{table:ablation_tokens}
    \vspace{-3mm}
    \scalebox{0.95}{
     \begin{tabular}{ c c c} 
    \toprule
     Embedding & 1-shot  & 5-shot \\ 
    \toprule
    { \textcircled{\small 1}}  cls  & ${{71.26\pm0.17}}$ & ${{88.03\pm0.10}}$  \\
    { \textcircled{\small 2}} avg pool  & ${{71.65\pm0.19}}$ & ${{86.40\pm0.11}}$  \\
    { \textcircled{\small 3}} weighted avg pool  & ${{71.83\pm0.19}}$ & ${{86.38\pm0.11}}$  \\
    { \textcircled{\small 1}} + { \textcircled{\small 2}} & ${{73.83\pm0.18}}$ & ${{88.58\pm0.10}}$  \\
    \rowcolor{lightblue}
    { \textcircled{\small 1}} + { \textcircled{\small 3}} & ${{74.28\pm0.18}}$  & ${{88.82\pm0.09}}$  \\
    { \textcircled{\small 2}} + { \textcircled{\small 3}} & ${{71.86\pm0.19}}$ & ${{86.47\pm0.11}}$  \\
    { \textcircled{\small 1}} + { \textcircled{\small 2}} + { \textcircled{\small 3}} & ${{73.80\pm0.18}}$ & ${{88.26\pm0.10}}$  \\ 
    \bottomrule
    \end{tabular}}
    \end{center}
\vspace{-8mm}
\end{table}
\normalsize

\section{Conclusion}

In this work, we propose a novel supervised knowledge distillation framework (SMKD) for few-shot Transformers, which extends the self-supervised masked knowledge distillation framework into the supervised setting. With our design of supervised-contrastive losses, we incorporate supervision into both class and patch-level knowledge distillation while still enjoy the benefits of not needing large batch size and negative samples. Evaluation results together with ablation studies demonstrate the superiority of our method given its simple design compared with contemporary works. 
Our two-stage training is a special case of curriculum
learning from the easy samples to the hard ones. We can
unify the learning objectives of self-supervised learning and
supervised contrastive learning, using a carefully designed
curriculum learning strategy for future works.
We hope our work can bridge the gap between self-supervised and supervised knowledge distillation, and inspire more works on supervised few-shot learning methods.


\section{Acknowledgements}
{
\noindent This research is sponsored by Air Force Research Laboratory (AFRL) under agreement number FA8750-19-1-1000. 
The U.S. Government is authorized to reproduce and distribute reprints for Government purposes notwithstanding any copyright notation therein. 
The views and conclusions contained herein are those of the authors and should not be interpreted as necessarily representing the official policies or endorsements, either expressed or implied, of Air Force Laboratory, DARPA or the U.S. Government.
}

\newpage
{\small
\bibliographystyle{ieee_fullname}

}

\newpage

\onecolumn\section{Appendix}

\subsection{Implementation Details for Model Training}\label{appsec:implementation_details}

\paragraph{Self-supervised pretraining.} We pre-train our Vision Transformer backbone and projection head following the same pipeline in iBOT \cite{zhou2021ibot}. Most of the hyper-parameter settings are kept unchanged without tuning. Vit-Small, which has $\sim$21M parameters is used as our default architecture. Our default patch size is set as 16. For the student network, the \texttt{[cls]} token output and \texttt{[patch]} tokens output share the same projection head. This head-sharing strategy is also applied to the teacher network. For both networks, we set the output dimension of projection heads as 8192. We linearly warm up the learning rate for 10 epochs to its base value of 5e-4, then use cosine schedule to decay it to 1e-5. Cosine schedule is also used for weight decay from 0.04 to 0.4. Besides, we use the multi-crop strategy \cite{caron2020unsupervised} with 2 global crops (224$\times$224) and 10 local crops (96$\times$96), with scale range $(0.4, 1.0)$ and $(0.05, 0.4)$ respectively. We found that allowing knowledge distillation between global and local crops from intra-class images harms the performance, which is consistent with \cite{zhou2021ibot}. Therefore, local crops here are only used for self-distillation with global crops from the same image. Furthermore, we apply blockwise masking on global crops sent into the student network, with a masking ratio uniformly sampled from $[0,1, 0.5]$ with probability 0.5, and 0 with probability 0.5. Ablation of different masking strategies is given in \cref{appsec:additional_ablations}. Our batch size is set as 640 (batch size per GPU equal to 80). \emph{mini}-ImageNet and \emph{tiered}-ImageNet are pre-trained for 1200 epochs, and CIFAR-FS and FC100 are pre-trained for 900 epochs. All models are trained on 8 Nvidia RTX 3090 GPUs.

\vspace{-3mm}
\paragraph{Supervised knowledge distillation.} After finishing the pretraining stage, we train the model with our supervised-contrastive loss. The best evaluation accuracy on the validation set can usually be achieved within 60 epochs of training. We use the same set of hyper-parameters as the first pretraining stage without further tuning. Ablation of the scaling parameter $\lambda$, which controls the relative size of $\mathbf{\mathcal{L}_{\texttt{[patch]}}}$ and $\mathbf{\mathcal{L}_{\texttt{[cls]}}}$ is given in \cref{appsec:additional_ablations}.

\vspace{2mm}
\subsection{Few-shot Evaluation Results}\label{appsec:additional_results}

We present few-shot evaluation results with more methods on the four benchmark datasets here in Table \ref{appendix:full_results_imagenet} and \ref{appendix:full_results_cifar}. ViT-based methods are better than the traditional CNN-based methods in general. The ranking of our method remains unchanged. \vspace{-3mm}
\begin{table*}[h!]
\begin{center}
\setlength{\tabcolsep}{0.7em}
\renewcommand{\arraystretch}{1.0}
\caption{\small{\textbf{More comprehensive few-shot evaluation results on mini-ImageNet and tiered-ImageNet.} Top three methods are colored in {\color{red} red}, {\color{blue} blue} and {\color{green} green} respectively.}}
\label{appendix:full_results_imagenet}
\vspace{-3mm}
\scalebox{0.97}{
\begin{tabular}{ccccc|cc}
\hlineB{3}
\multirow{2}{*}{Method} & \multirow{2}{*}{Backbone} & \multirow{2}{*}{\#Params} & \multicolumn{2}{c|}{miniImageNet,5-way} & \multicolumn{2}{c}{tieredImageNet,5-way} \\ \cline{4-7} 
 & & & 1-shot & 5-shot & 1-shot & 5-shot  \\ \hline
 DeepEMD~\cite{Zhang_2020_CVPR} & \emph{ResNet-12} & 12.4M & ${{65.91\pm0.82}}$  & ${{82.41\pm0.56}}$ & ${{71.16\pm0.87}}$ & ${{86.03\pm0.58}}$ \\
 IE~\cite{Rizve_2021_CVPR} & \emph{ResNet-12} & 12.4M &  ${{67.28\pm0.80}}$  & ${{84.78\pm0.52}}$ & ${{72.21\pm0.90}}$ & ${{87.08\pm0.58}}$ \\
 BML~\cite{zhou2021binocular} & \emph{ResNet-12}  & 12.4M & ${{67.04\pm0.63}}$  & ${{83.63\pm0.29}}$ & ${{68.99\pm0.50}}$ & ${{85.49\pm0.34}}$ \\
 PAL~\cite{ma2021partner} & \emph{ResNet-12} & 12.4M & ${{69.37\pm0.64}}$  & ${{84.40\pm0.44}}$ & ${{72.25\pm0.72}}$ & ${{86.95\pm0.47}}$ \\
 TPMN~\cite{wu2021task} & \emph{ResNet-12} & 12.4M & ${{67.64\pm0.63}}$  & ${{83.44\pm0.43}}$ & ${{72.24\pm0.70}}$ & ${{86.55\pm0.63}}$ \\
 MN+MC\cite{zhang2021meta} & \emph{ResNet-12} & 12.4M  & ${{67.14\pm0.80}}$  & ${{83.82\pm0.51}}$ & ${{74.58\pm0.88}}$ & ${{86.73\pm0.61}}$ \\
 DC~\cite{yang2021free} & \emph{ResNet-12} & 12.4M & ${{68.57\pm0.55}}$  & ${{82.88\pm0.42}}$ & ${{78.19\pm0.25}}$ & ${{89.90\pm0.41}}$ \\
 MELR~\cite{fei2020melr} & \emph{ResNet-12} & 12.4M & ${{67.40\pm0.43}}$  & ${{83.40\pm0.28}}$ & ${{72.14\pm0.51}}$ & ${{87.01\pm0.35}}$ \\
 COSOC~\cite{luo2021rectifying} & \emph{ResNet-12} & 12.4M & ${{69.28\pm0.49}}$  & ${{85.16\pm0.42}}$ & ${{73.57\pm0.43}}$ & ${{87.57\pm0.10}}$ \\
 CSEI~\cite{li2021learning} & \emph{ResNet-12}  & 12.4M & ${{68.94\pm0.28}}$  & ${{85.07\pm0.50}}$ & ${{73.76\pm0.32}}$ & ${{87.83\pm0.59}}$ \\    
 CNL~\cite{zhao2021looking} & \emph{ResNet-12}  & 12.4M & ${{67.96\pm0.98}}$  & ${{83.36\pm0.51}}$ & ${{73.42\pm0.95}}$ & ${{87.72\pm0.75}}$ \\   
 \hline
 FEAT~\cite{ye2020few} & \emph{WRN-28-10} & 36.5M & ${{65.10\pm0.20}}$  & ${{81.11\pm0.14}}$ & ${{70.41\pm0.23}}$ & ${{84.38\pm0.16}}$ \\
 Meta-QDA~\cite{zhang2021shallow} & \emph{WRN-28-10} & 36.5M & ${{67.38\pm0.55}}$  & ${{84.27\pm0.75}}$ & ${{74.29\pm0.66}}$ & ${{89.41\pm0.77}}$ \\
 OM~\cite{qi2021transductive} & \emph{WRN-28-10} & 36.5M & ${{66.78\pm0.30}}$  & ${{85.29\pm0.41}}$ & ${{71.54\pm0.29}}$ & ${{87.79\pm0.46}}$ \\
 \hline
 SUN~\cite{dong2022self} & \emph{ViT} & 12.5M & ${{67.80\pm0.45}}$  & ${{83.25\pm0.30}}$ & ${{72.99\pm0.50}}$ & ${{86.74\pm0.33}}$ \\ 
 FewTURE~\cite{hiller2022rethinking} & \emph{ViT-S} & 21.0M & ${{68.02\pm0.88}}$  & ${{84.51\pm0.53}}$ & ${{72.96\pm0.92}}$ & ${{86.43\pm0.67}}$ \\ 
 FewTURE~\cite{hiller2022rethinking} & \emph{Swin-Tiny} & 29.0M & ${{72.40\pm0.78}}$  & ${{86.38\pm0.49}}$ & ${{76.32\pm0.87}}$ & ${{89.96\pm0.55}}$ \\ 
 HCT (Prototype)~\cite{he2022hct} & 3$\times$\emph{ViT-S} & 63.0M & ${\color{blue}{74.74\pm0.17}}$ & ${{85.66\pm0.10}}$ & ${\color{blue}{79.67\pm0.20}}$ & ${{89.27\pm0.13}}$ \\  
 HCT (Classifier)~\cite{he2022hct} & 3$\times$\emph{ViT-S} & 63.0M  & ${\color{green}{74.62\pm0.20}}$ & ${\color{blue}{89.19\pm0.13}}$ & ${\color{green}{79.57\pm0.20}}$ & ${\color{red}{91.72\pm0.11}}$ \\  
 \hline
 Ours (Prototype)  & \emph{ViT-S} & 21.0M & ${{74.28\pm0.18}}$  & ${{88.82\pm0.09}}$  &    ${{78.83\pm0.20}}$     &  ${{91.02\pm0.12}}$       \\
Ours (Classifier)  & \emph{ViT-S} & 21.0M & ${{74.10\pm0.17}}$  & ${\color{green}{88.89\pm0.09}}$  &    ${{78.81\pm0.21}}$     &  ${\color{green}{91.21\pm0.11}}$       \\ 
Ours + HCT~\cite{he2022hct}  & 3$\times$\emph{ViT-S} & 63.0M & ${\color{red}{75.32\pm0.18}}$  & ${\color{red}{89.57\pm0.09}}$  &    ${\color{red}{79.74\pm0.20}}$     &  ${\color{blue}{91.68\pm0.11}}$       \\ 
\hlineB{3}
\end{tabular}}
\end{center}
\vspace{-2mm}
\end{table*}

\begin{table*}[h]
\begin{center}
\setlength{\tabcolsep}{0.7em}
\renewcommand{\arraystretch}{1.0}
\caption{\small{\textbf{More comprehensive few-shot evaluation results on CIFAR-FS and FC100}.} Top three methods are colored in {\color{red} red}, {\color{blue} blue} and {\color{green} green} respectively.}
\label{appendix:full_results_cifar}
\vspace{-3mm}
\scalebox{0.97}{
\begin{tabular}{ccccc|cc}
\hlineB{3}
\multirow{2}{*}{Method} & \multirow{2}{*}{Backbone} & \multirow{2}{*}{\#Params} & \multicolumn{2}{c|}{CIFAR-FS,5-way} & \multicolumn{2}{c}{FC100,5-way} \\ \cline{4-7} 
 & & & 1-shot & 5-shot & 1-shot & 5-shot  \\ \hline
DSN-MR~\cite{simon2020adaptive} & \emph{ResNet-12} & 12.4M & ${{75.60\pm0.90}}$  & ${{86.20\pm0.60}}$ & - & - \\ 
BML~\cite{zhou2021binocular} & \emph{ResNet-12} & 12.4M & ${{73.45\pm0.47}}$  & ${{88.04\pm0.33}}$ & ${{45.00\pm0.41}}$  & ${{63.03\pm0.41}}$ \\
IE~\cite{Rizve_2021_CVPR} & \emph{ResNet-12} & 12.4M & ${{77.87\pm0.85}}$  & ${{89.74\pm0.57}}$ & ${{47.76\pm0.77}}$  & ${{65.30\pm0.76}}$ \\
PAL~\cite{ma2021partner} & \emph{ResNet-12} & 12.4M & ${{77.10\pm0.70}}$  & ${{88.00\pm0.50}}$ & ${{47.20\pm0.60}}$  & ${{64.00\pm0.60}}$ \\
TPMN~\cite{wu2021task} & \emph{ResNet-12} & 12.4M  & ${{75.50\pm0.90}}$  & ${{87.20\pm0.60}}$ & ${{46.93\pm0.71}}$  & ${{63.26\pm0.74}}$ \\
MN+MC~\cite{zhang2021meta} & \emph{ResNet-12} & 12.4M & ${{74.63\pm0.91}}$  & ${{86.45\pm0.59}}$ & ${{46.40\pm0.81}}$  & ${{61.33\pm0.71}}$   \\
RENet~\cite{kang2021relational} & \emph{ResNet-12} & 12.4M & ${{74.51\pm0.46}}$  & ${{86.60\pm0.32}}$ & - & -  \\
ConstellationNet~\cite{xu2021attentional} & \emph{ResNet-12} & 12.4M & ${{75.40\pm0.20}}$  & ${{86.80\pm0.20}}$ & ${{43.80\pm0.20}}$  & ${{59.70\pm0.20}}$ \\
ALFA+MeTAL~\cite{baik2021meta} & \emph{ResNet-12} & 12.4M & - & - & ${{44.54\pm0.50}}$  & ${{58.44\pm0.42}}$  \\
MixtFSL~\cite{afrasiyabi2021mixture} & \emph{ResNet-12} & 12.4M & - & - & ${{41.50\pm0.67}}$  & ${{58.39\pm0.62}}$  \\
\hline
CC+rot~\cite{gidaris2019boosting} & \emph{WRN-28-10} & 36.5M & ${{73.62\pm0.31}}$  & ${{86.05\pm0.22}}$ & - & -  \\
PSST~\cite{chen2021pareto} & \emph{WRN-28-10} & 36.5M & ${{77.02\pm0.38}}$  & ${{88.45\pm0.35}}$ & - & -  \\
Meta-QDA~\cite{zhang2021shallow} & \emph{WRN-28-10} & 36.5M &  ${{75.95\pm0.59}}$  & ${{88.72\pm0.79}}$ & - & -  \\
\hline
SUN~\cite{dong2022self} & \emph{ViT} & 12.5M & ${{78.37\pm0.46}}$  & ${{88.84\pm0.32}}$ & - & -  \\   
FewTURE~\cite{hiller2022rethinking} & \emph{ViT-S} & 21.0M & ${{76.10\pm0.88}}$  & ${{86.14\pm0.64}}$ & ${{46.20\pm0.79}}$  & ${{63.14\pm0.73}}$ \\  
FewTURE~\cite{hiller2022rethinking} & \emph{Swin-Tiny} & 29.0M & ${{77.76\pm0.81}}$  & ${{88.90\pm0.59}}$ & ${{47.68\pm0.78}}$  & ${{63.81\pm0.75}}$  \\  
HCT (Prototype)~\cite{he2022hct} & 3$\times$\emph{ViT-S} & 63.0M & ${\color{green}{78.89\pm0.18}}$ & ${{87.73\pm0.11}}$ & ${\color{green}{48.27\pm0.15}}$ & ${{61.49\pm0.15}}$ \\  
HCT (Classifier)~\cite{he2022hct} & 3$\times$\emph{ViT-S} & 63.0M & ${{78.88\pm0.18}}$ & ${\color{green}{90.50\pm0.09}}$ & ${{48.15\pm0.16}}$ & ${\color{green}{66.42\pm0.16}}$ \\  
\hline
Ours (Prototype) & \emph{ViT-S} & 21M & ${\color{red}{80.08\pm0.18}}$   &  ${\color{blue}{90.63\pm0.13}}$   & ${\color{red}{50.38\pm0.16}}$   &  ${\color{blue}{68.37\pm0.16}}$  \\
Ours (Classifier)  & \emph{ViT-S} & 21M & ${\color{blue}{79.82\pm0.18}}$  & ${\color{red}{90.91\pm0.13}}$  &    ${\color{blue}{50.28\pm0.16}}$     &  ${\color{red}{68.50\pm0.16}}$       \\ 
\hlineB{3}
\end{tabular}}
\end{center}
\vspace{-6mm}
\end{table*}

\vspace{10mm}
\subsection{Additional Ablation Studies}\label{appsec:additional_ablations}

\paragraph{Why $\mathcal{L}_{[cls]} + \mathcal{L}_{\text{MIM}}$ in stage 1?} Our insight is that the \texttt{[cls]} tokens in global loss have better high-level semantics, but often disregard the rich local structures. While the MIM loss $\mathcal{L}_{\text{MIM}}$ constructed from \texttt{[patch]} tokens can remedy this problem, increase task difficulty, and work as strong data augmentations. In Table \ref{table:ablation_stage1}, we can find that using both losses in stage 1 gives the best results.

\vspace{-0mm}
\begin{table}[h!]
\begin{center}
\setlength{\tabcolsep}{0.7em}
\renewcommand{\arraystretch}{1.0}
\caption{\small{Ablation of SSL tasks in stage 1 on {\sl mini}-ImageNet}.} 
\label{table:ablation_stage1}
\vspace{-2.5mm}
\scalebox{0.99}{
\begin{tabular}{c@{\hskip0.1in}c@{\hskip0.1in}c@{\hskip0.1in}c|cc}
\hlineB{3}
\multicolumn{4}{c|}{Stage1} & \multicolumn{2}{c}{Stage2: $\mathcal{L}_{\texttt{[cls]}}+\mathcal{L}_{\texttt{[patch]}}$} \\ \cline{1-6} 
$\mathcal{L}_{[cls]}$ & $\mathcal{L}_{\text{MIM}}$ & 1-shot   & 5-shot    & 1-shot   & 5-shot \\ \hline
\checkmark &  & 58.55 & 78.90  & 72.93 & 88.07 \\
 & \checkmark & 27.66 & 33.82 & 37.03 & 50.95  \\
\rowcolor{lightblue}
\checkmark & \checkmark & 60.93 & 80.38 & 74.28  & 88.82  \\ 
\hlineB{3}
\vspace{-10mm}
\end{tabular}}
\end{center}
\end{table}

\paragraph{Masking Strategies.} We use blockwise masking as our default in the main text. In Table \ref{table:ablation_masking_strategy}, we test random mask and no mask while keeping all other hyper-parameters unchanged. "Block Mask $\xrightarrow[]{}$ No Mask" represents self-supervised pretraining with blockwise masking, and supervised training with no mask. Using either a random mask or block mask can boost the classification accuracy in the first self-supervised pretraining stage, but their advantage over no mask decreases in the second supervised training stage. We choose blockwise masking as our default strategy since it balances 1 and 5-shot classification accuracy the best.

\begin{table*}[h]
\begin{center}
\setlength{\tabcolsep}{0.7em}
\renewcommand{\arraystretch}{1.0}
\caption{\small{Ablation over different masking strategy in self-supervised pretraning stage.}}
\label{table:ablation_masking_strategy}
\vspace{-2mm}
\begin{tabular}{ccc|cc}
\hlineB{3}
\multirow{2}{*}{Masking Strategy} &  \multicolumn{2}{c|}{Self-supervised Pre-train} & \multicolumn{2}{c}{Supervised Training} \\ \cline{2-5} 
& 1-shot   & 5-shot    & 1-shot   & 5-shot \\ \hline
No Mask & ${{59.15\pm0.17}}$ & ${{79.23\pm0.12}}$  & ${{73.94\pm0.17}}$ & ${{88.93\pm0.09}}$ \\
\rowcolor{lightblue}
Block Mask & ${{60.93\pm0.17}}$ & ${{80.38\pm0.12}}$ & ${{74.01\pm0.17}}$ &  ${{88.89\pm0.09}}$ \\
Random Mask & ${{60.94\pm0.18}}$ & ${{79.62\pm0.13}}$ & ${{74.07\pm0.18}}$ & ${{88.66\pm0.09}}$ \\ 
Block Mask $\xrightarrow[]{}$ No Mask & ${{60.93\pm0.17}}$ & ${{80.38\pm0.12}}$ & ${{73.44\pm0.17}}$ & ${{88.87\pm0.09}}$ \\ 
\hlineB{3}
\end{tabular}
\end{center}
\end{table*}

\paragraph{Scaling Parameter $\lambda$.} 
This parameter controls the relative importance of class-level and patch-level losses in our final loss: $\mathbf{\mathcal{L}} = \mathbf{\mathcal{L}_{[cls]}} + \lambda \mathbf{\mathcal{L}_{[patch]}}$. A relatively large value of $\lambda$ will put more focus on localization and less on high-level semantics. Here in Table \ref{table:ablation_loss_ratio}, we test different $\lambda$ values by keeping the base of $\mathbf{\mathcal{L}_{[cls]}}$ to 1 and scale $\mathbf{\mathcal{L}_{[patch]}}$. As we can see, the $\lambda$ parameter influences 1-shot classification accuracy more than 5-shot. We choose $\lambda=0.25$ as our default (which makes the ratio of $\mathbf{\mathcal{L}_{[patch]}}/\mathbf{\mathcal{L}_{[cls]}}$ roughly around 2) since it has best 1-shot performance and competitive 5-shot accuracy.

\vspace{-1mm}
\begin{table}[h!]
    \begin{center}
    \caption{\small{The influence of different ratio between $\mathbf{\mathcal{L}_{[cls]}}$ and $\mathbf{\mathcal{L}_{[patch]}}$. }}
    \label{table:ablation_loss_ratio}
    \vspace{-2mm}
    \scalebox{0.99}{
     \begin{tabular}{ c c c} 
    \toprule
     $\mathbf{\mathcal{L}_{[patch]}}/\mathbf{\mathcal{L}_{[cls]}}$ & 1-shot & 5-shot \\ [0.5ex]
    \toprule
     $\approx$4 ($\lambda=0.9$) & ${{73.45\pm0.17}}$ & ${{88.89\pm0.09}}$  \\
     \rowcolor{lightblue}
     $\approx$2 ($\lambda=0.45$)& ${{74.28\pm0.18}}$  & ${{88.82\pm0.09}}$  \\
     $\approx$1 ($\lambda=0.2$) &  ${{74.01\pm0.17}}$  & ${{88.89\pm0.09}}$ \\
     $\approx$0.5 ($\lambda=0.1$) &  ${{73.11\pm0.17}}$ & ${{88.44\pm0.09}}$  \\
    \bottomrule
    \end{tabular}}
    \end{center}
\end{table}
\normalsize

\paragraph{Weighting Parameter $\omega_{k^+}$ in $\mathbf{\mathcal{L}_{[patch]}}$.} This parameter in Eq.(\ref{eq:sc_patch}) gives weights to each component of our patch-level contrastive loss $\mathbf{\mathcal{L}_{[patch]}}$. We set $\omega_{k^+}=1/N$ in our main text due to its simplicity. In Table \ref{table:ablation_patch_loss_design}, we compare it \emph{(Simple Avg)} with another variant \emph{(Self-Attention Weighted Avg)}, which uses the averaged self-attention weights over attention heads of the \texttt{[cls]} token with all \texttt{[patch]} tokens in the last attention layer of teacher network to aggregate pairwise patch matching losses. As found in \cite{caron2021emerging}, the self-attention of ViTs is good at capturing foreground regions. So we use it here as a way to highlight foreground objects and to attenuate irrelevant background information. Our default simple average outperforms this variant on both 1 and 5-shot classification accuracies. One explanation is as follows. If the foreground objects of two intra-class images differ a lot, then \emph{Self-Attention Weighted Avg} tends to minimize $\mathbf{\mathcal{L}_{[patch]}}$ by decreasing the weights of the losses associated with the patches covering these foreground objects, which makes our model deviate from optimum.

\begin{table}[h!]
    \begin{center}
    \caption{\small{Different weighting schemes of patch-level supervised-contrastive loss. }}
    \label{table:ablation_patch_loss_design}
    \vspace{-2mm}
    \scalebox{0.99}{
     \begin{tabular}{ c c c} 
    \toprule
     Weighting Scheme & 1-shot & 5-shot \\ [0.5ex]
    \toprule
         \rowcolor{lightblue}
    Simple Avg & ${{74.28\pm0.18}}$ & ${{88.82\pm0.09}}$  \\
    Self-Attention Weighted Avg & ${{74.11\pm0.18}}$  & ${{88.52\pm0.10}}$  \\
    \bottomrule
    \end{tabular}}
    \end{center}
\end{table}
\normalsize

\paragraph{Comparison with smaller backbones:} To make ViT-S ($\sim$ 21M parameters) comparable with ResNet-12 ({$\sim$ 12M} parameters), we trim by half either its embedding dimension ($d_{\text{embed}}$) or the number of attention heads (\#heads). 
From Table \ref{table:ablation_half_parameter}, trimming \#heads by half only results in little drop in accuracy, which still outperforms the best method with ResNet-12 backbones. The training speed also increases by 10\% with fewer \#heads. Given this result, our comparison now becomes complete: our method outperforms both shallow (ResNet-12) and deep (WRN-28-10) CNN-based backbones, as well as ViTs with the same (see Table \ref{table:comparison_same_backbone}) or more (see Table \ref{table:our_best} \& Table \ref{table:comparison_training_clock_time}) parameters.

\begin{table}[h!]
\begin{center}
\setlength{\tabcolsep}{0.7em}
\renewcommand{\arraystretch}{1.0}
\caption{\small{ViT-S with similar size as ResNet-12 on {\sl mini}-ImageNet}}
\label{table:ablation_half_parameter}
\vspace{-2mm}
\scalebox{0.99}{
\begin{tabular}{c@{\hskip0.06in}c@{\hskip0.06in}c@{\hskip0.06in}c@{\hskip0.1in}c@{\hskip0.08in}c|c@{\hskip0.08in}c}
\hlineB{3}
\multirow{2}{*}{Backbone} & \multirow{2}{*}{$d_{\text{embed}}$} & \multirow{2}{*}{\#heads}  & \multirow{2}{*}{\#param} & \multicolumn{2}{c|}{Stage1} & \multicolumn{2}{c}{Stage2} \\ \cline{5-8} 
 & & & & 1-shot   & 5-shot    & 1-shot   & 5-shot \\ \hline
 ViT-S & \textbf{192} & 6  & 11M & 60.70 & 79.56  & 71.14 & 87.12 \\
ViT-S & 384 & \textbf{3} & 11M & 62.12 & 81.27 & \textbf{72.70} & \textbf{87.90} \\
ViT-S & 384 & 6 & 21M & 60.93  & 80.38  & 74.28  & 88.82 \\ 
\hline
 ResNet12 & - & - & 12M  & -  & - & 69.37  & 85.16  \\ 
WRN-28-10 & - & - & 36M  & -  & - & 67.38  & 85.29  \\ 
\hlineB{3}
\vspace{-14mm}
\end{tabular}}
\end{center}
\end{table}

\newpage
\subsection{Visualizations}\label{appsec:visualizations}

\begin{figure}[h!]
\begin{center}
\includegraphics[width=15cm, height=1.9cm]{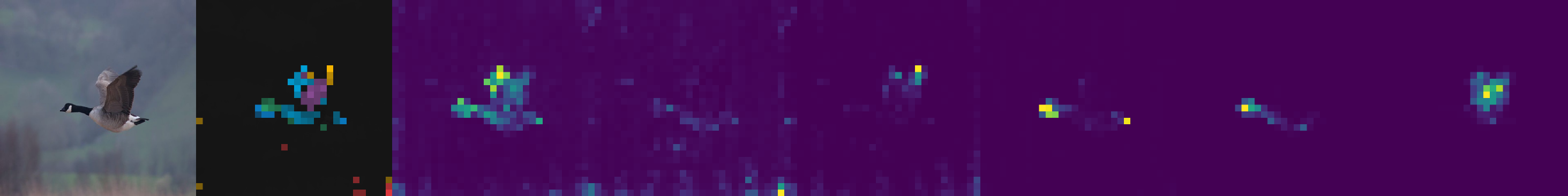}
\vspace{0mm}
\includegraphics[width=15cm, height=1.9cm]{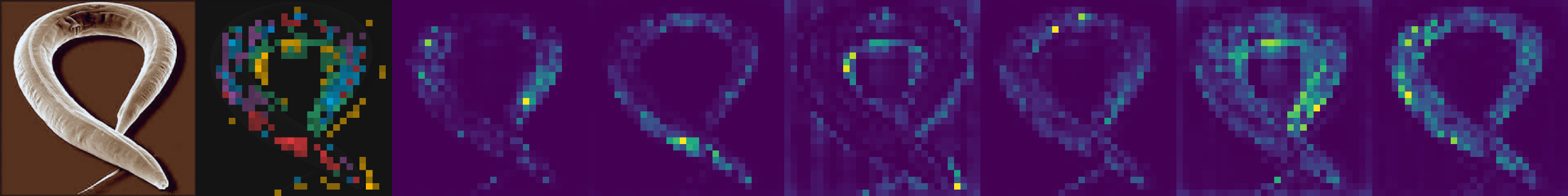}
\vspace{0mm}
\includegraphics[width=15cm, height=1.9cm]{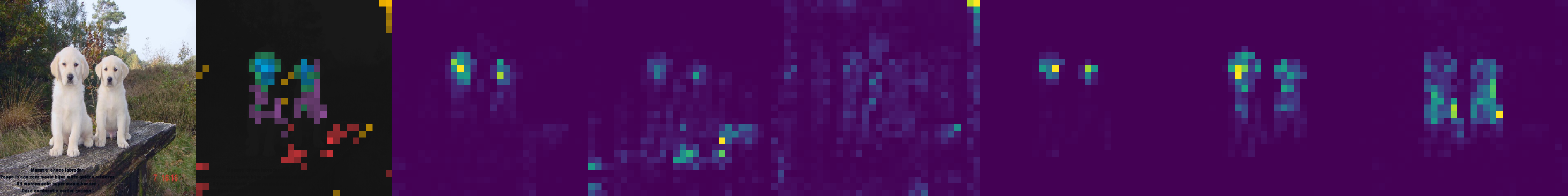}
\vspace{0mm}
\includegraphics[width=15cm, height=1.9cm]{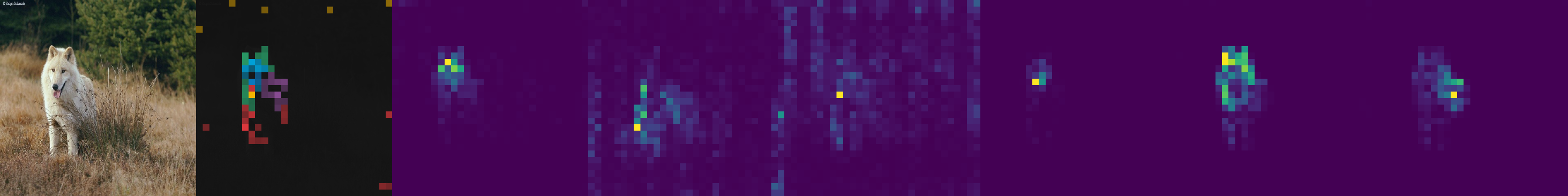}
\vspace{0mm}
\includegraphics[width=15cm, height=1.9cm]{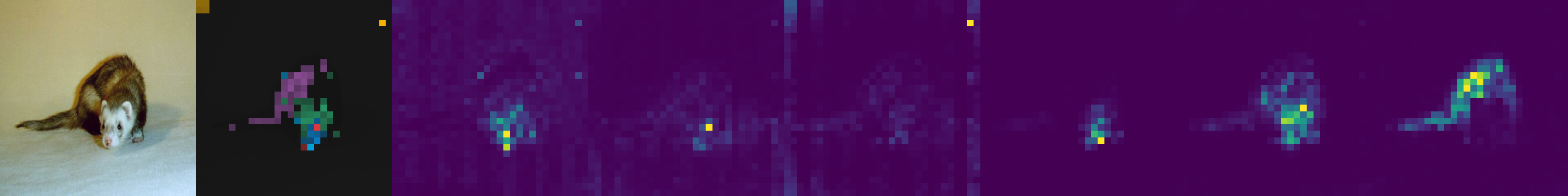}
\vspace{0mm}
\includegraphics[width=15cm, height=1.9cm]{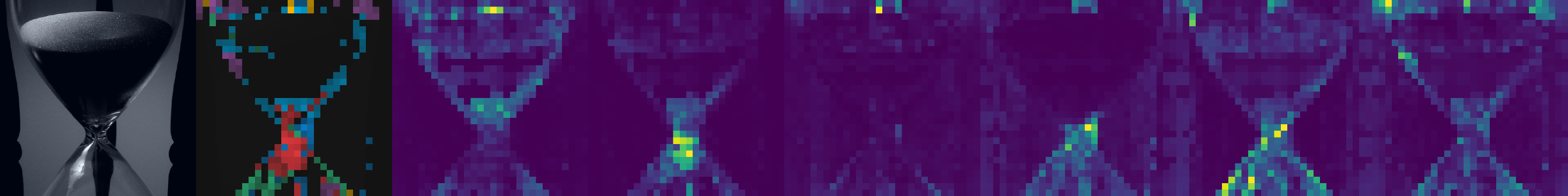}
\vspace{0mm}
\includegraphics[width=15cm, height=1.9cm]{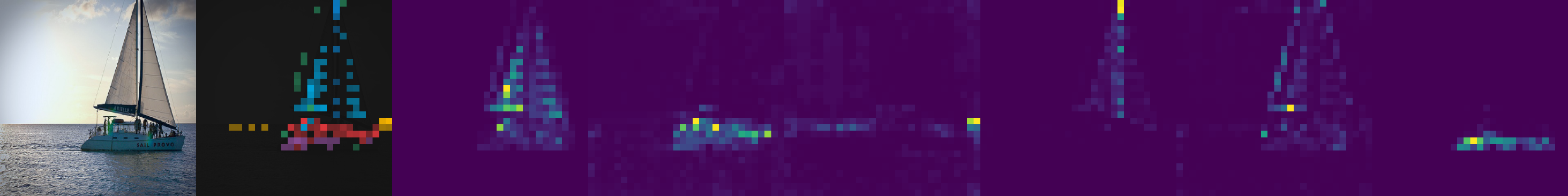}
\vspace{0mm}
\includegraphics[width=15cm, height=1.9cm]{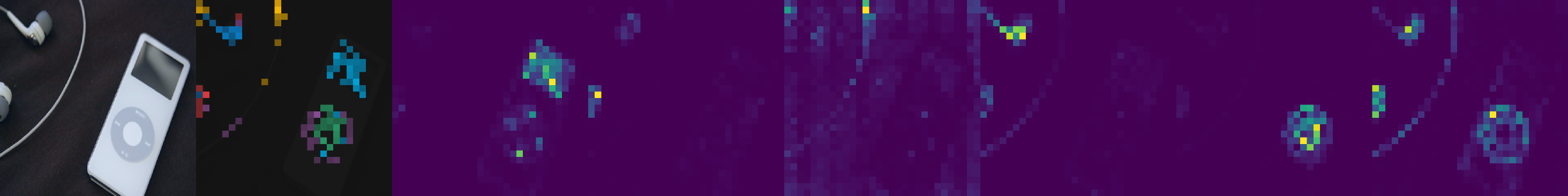}
\vspace{0mm}
\includegraphics[width=15cm, height=1.9cm]{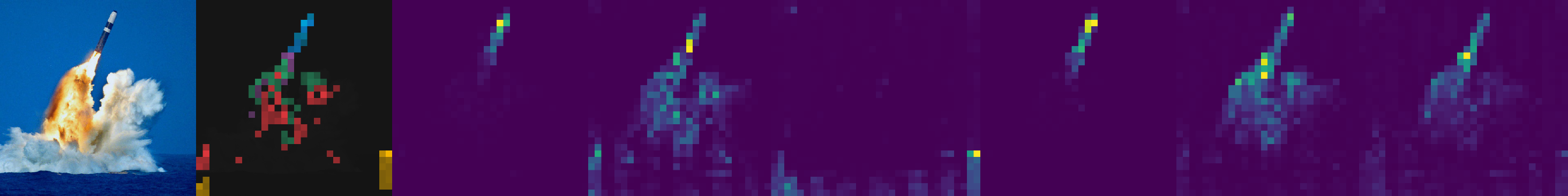}
\vspace{0mm}
\includegraphics[width=15cm, height=1.9cm]{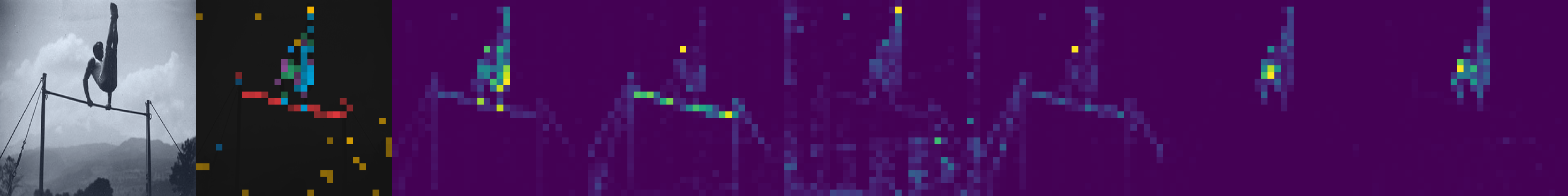}
\vspace{-2mm}
\caption{\small{\textbf{Visualization of multi-head self-attention maps}. The self-attention of the $\texttt{[cls]}$ tokens with different heads in the last attention layer of ViT are visualized in different colors in the second column. The last six columns visualize each attention head. Images are from the test set of \emph{mini}-ImageNet.}}
\label{fig:attention_maps_appendix}
\vspace{-2mm}
\end{center}
\end{figure}

\begin{figure}[h!]  
    \includegraphics[width=.99\linewidth]{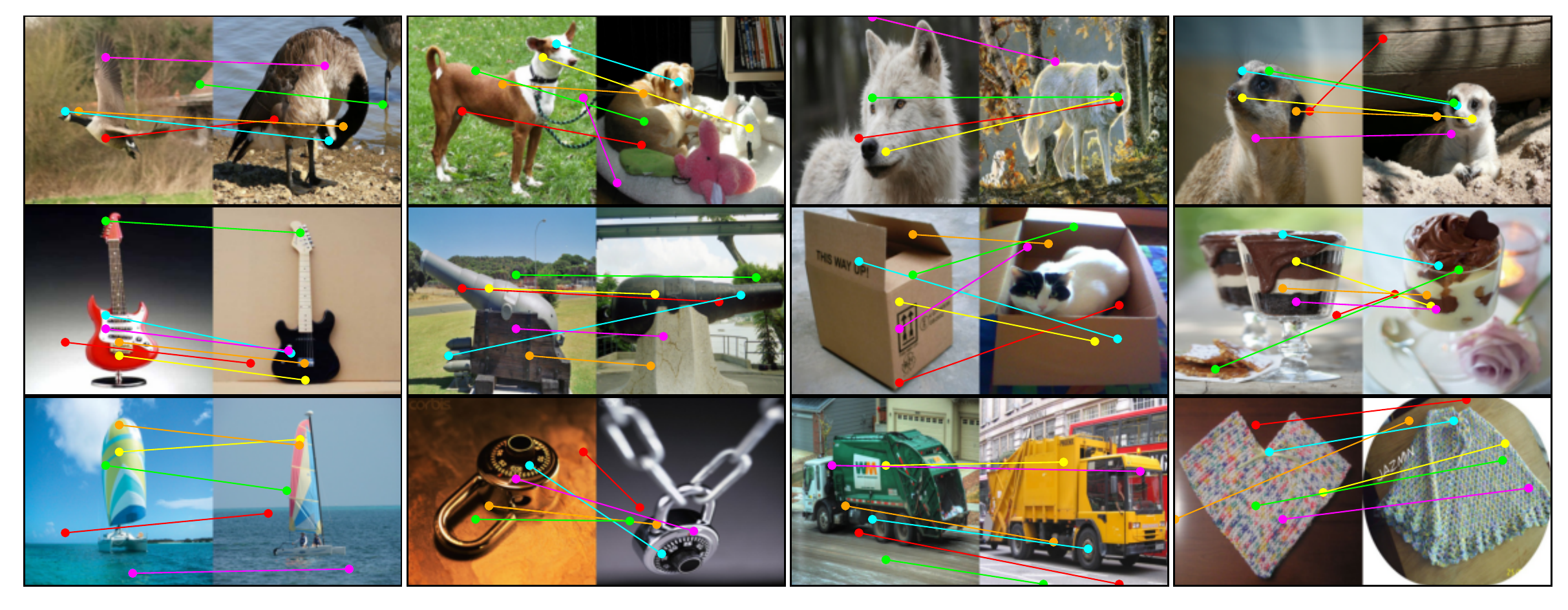}
    \vspace{-4mm}
    \caption{\small{\textbf{Visualization of dense correspondence}. We use the patches with the highest self-attention of the $\texttt{[cls]}$ token on each attention head (6 in total) of the last layer of ViT-S as queries. Best-matched patches with the highest similarities are connected with lines. Images are from the validation and testing set of \emph{mini}-ImageNet.}}
\label{fig:dense_match_appendix}
\vspace{-2mm}
\end{figure}

We visualize more self-attention maps and dense correspondence in Fig. \ref{fig:attention_maps_appendix} and Fig. \ref{fig:dense_match_appendix}.

\pagebreak

\end{document}